\documentclass[draft]{agujournal2019}
\usepackage{url} 
\usepackage[inline]{trackchanges} 
\usepackage{soul}
\usepackage{amsmath}
\usepackage{amssymb}

\usepackage{algpseudocode}
\usepackage{algorithm}
\usepackage{booktabs}
\usepackage{tabularx}
\usepackage{makecell}
\usepackage{xcolor}

\usepackage{subcaption}  

\usepackage{array}

\renewcommand*{\arraystretch}{1.1}
\draftfalse

\journalname{JGR: Machine Learning and Computation}

\begin{document}

%
%


\title{Conditional Diffusion Models\\for Global Precipitation Map Inpainting}

%
%




\authors{Daiko Kishikawa\affil{1}, Yuka Muto\affil{1}, Shunji Kotsuki\affil{1,2,3,4}}

\affiliation{1}{Center for Environmental Remote Sensing, Chiba University}
\affiliation{2}{Institute for Advanced Academic Research, Chiba University}
\affiliation{3}{Research Institute of Disaster Medicine, Chiba University}
\affiliation{4}{Data Assimilation Research Team, RIKEN Center for Computational Science}

\affiliation{1,2,3}{1-33 Yayoi-cho, Inage-ku, Chiba-shi, Chiba, Japan, 263-8522}
\affiliation{4}{7-1-26 Minatojima-minami-machi, Chuo-ku, Kobe-Shi, Hyogo, Japanm 650-0047}

\correspondingauthor{Daiko Kishikawa}{d.kskw@chiba-u.jp}
\correspondingauthor{Shunji Kotsuki}{shunji.kotsuki@chiba-u.jp}

\begin{keypoints}
\item High-fidelity global precipitation maps are in high demand, but polar-orbiting satellite cause spatio-temporal gaps.
\item We propose a diffusion model-based inpainting method using a 3D U-Net to obtain spatio-temporal consistency.
\item Our method achieves higher spatio-temporal consistency than conventional and deep learning approaches in filling unobserved regions.
\end{keypoints}

\begin{abstract}
    Incomplete satellite-based precipitation presents a significant challenge in global monitoring. For example, the Global Satellite Mapping of Precipitation (GSMaP) from JAXA suffers from substantial missing regions due to the orbital characteristics of satellites that have microwave sensors, and its current interpolation methods often result in spatial discontinuities. In this study, we formulate the completion of the precipitation map as a video inpainting task and propose a machine learning approach based on conditional diffusion models. Our method employs a 3D U-Net with a 3D condition encoder to reconstruct complete precipitation maps by leveraging spatio-temporal information from infrared images, latitude-longitude grids, and physical time inputs. Training was carried out on ERA5 hourly precipitation data from 2020 to 2023. We generated a pseudo-GSMaP dataset by randomly applying GSMaP masks to ERA maps. Performance was evaluated for the calendar year 2024, and our approach produces more spatio-temporally consistent inpainted precipitation maps compared to conventional methods. These results indicate the potential to improve global precipitation monitoring using the conditional diffusion models.
\end{abstract}

\section*{Plain Language Summary}
Current global precipitation maps use data from multiple satellites to estimate rainfall amounts. However, polar-orbiting satellites cannot simultaneously observe the entire Earth, which means that large portions of the precipitation maps lack observations. Current methods fill in these missing areas by estimating values from the previous and next hours' observations. Unfortunately, these methods mainly focus on time-based consistency and sometimes generate predictions that are not smoothly connected to observed areas, resulting in noticeable gaps.

To overcome this issue, we developed a new machine learning approach. Instead of relying on manually designed formulas, our method treats precipitation map sequences as video-like data and repairs missing sections by using information from both nearby time frames and the surrounding regions with additional information. We built a model that generates three-dimensional data and learns to complete precipitation maps using additional inputs such as infrared images, geographic coordinates, and time information.

We trained our model to complete hourly precipitation data from 2020 to 2023 by simulating the missing regions using actual satellite observations. The results show that our model produces smoother, more consistent precipitation maps that effectively fill in the missing areas compared to conventional methods.

\section{Introduction}

\subsection{Background and Motivation}
In recent years, high-accuracy and high-temporal-resolution global precipitation maps has become essential for mitigating damage from extreme weather, supporting water management activities (e.g., dam operations), and informing public policy decisions at both governmental and local levels. Thus, improving global precipitation maps have become an increasingly important topic in current research.

One notable example is the Global Satellite Mapping of Precipitation (GSMaP) \cite{kubota2020global} produced by the Japan Aerospace Exploration Agency (JAXA). GSMaP uses microwave observations from multiple polar-orbiting satellites to estimate global precipitation. However, due to orbital limitations, it is practically impossible to observe the entire globe simultaneously.
Another example is the Integrated Multi-satellite Retrievals for GPM (IMERG) \cite{huffman2014imerg} produced by the National Aeronautics and Space Administration (NASA). IMERG also relies on microwave observations from polar-orbiting satellites, resulting in similar spatio-temporal gaps due to orbital constraints.

To make these products operationally usable, large spatio-temporal gaps must be filled. For that purpose, conventional inpainting methods are applied using transformation formulas based on adjacent temporal observations to achieve temporal consistency both in GSMaP and IMERG. 
However, spatial consistency is not prioritized, often resulting in inconsistencies between observed and inpainted regions. Hence, achieving both spatial and temporal consistency remains a significant challenge in satellite-based global precipitation mapping.

\subsection{Conventional and Deep Learning Approaches}
Over the past decades, a variety of methods --- from classical statistical approaches to modern deep learning techniques --- have been proposed to tackle this inpainting problem. Conventional methods can be divided into two categories: spatial methods and data assimilation methods.

Conventional techniques for inpainting precipitation maps include methods such as Inverse Distance Weighting (IDW) and Kriging \cite{cressie1990origins}. These methods estimate values in unobserved regions by leveraging data from neighboring locations. Advanced Kriging-based techniques, such as co-kriging and kriging with external drift, have demonstrated superior predictive performance compared to simpler approaches like IDW \cite{caloiero2021comparative}. However, purely spatial methods often produce overly smooth and physically unrealistic precipitation fields \cite{geiss2021inpainting}. Other approaches based on Markov Random Fields (MRF) \cite{cheng2014cloud} and the Total Variation (TV) method \cite{yang2009fast} rely on specific assumptions about the underlying image statistics, which may limit their effectiveness for large unobserved regions. 

Data assimilation represents another approach to produce precipitation analyses by combining predictions from numerical weather prediction (NWP) models with observations from multiple sensors, including satellite precipitation data \cite{kotsuki2017assimilating}. However, data assimilation approaches are limited by the horizontal resolution of NWP models, making it difficult to generate high-resolution precipitation data.

Recent advances in deep learning have provided a promising alternative. Convolutional Neural Networks (CNNs) and encoder--decoder architectures, such as U-Net, can learn the structural characteristics of precipitation fields through a purely data-driven approach. For example, Geiss and Hardin \cite{geiss2021inpainting} demonstrated that a CNN trained on radar images could effectively inpaint missing regions that radar could not scan. Both pure CNN models and Generative Adversarial Networks (GANs) \cite{goodfellow2014generative} have been reported to outperform conventional methods such as linear interpolation and Laplace equation-based approaches.
In particular, GAN-based inpainting combines adversarial loss with pixel-level error measures to enable high-fidelity reconstruction. Furthermore, recent models based on U-Net \cite{ronneberger2015u} have been shown to substantially reduce errors in radar blind zones (i.e., near-surface gaps in satellite radar observations), with the incorporation of meteorological features as conditional inputs reducing the mean absolute error (MAE) by an additional 6\% \cite{king2024development}.

\subsection{Our Diffusion Model Approach}
Our main contributions are as follows:
\begin{itemize}
  \item Applying diffusion models to inpainting problem of satellite-derived global precipitation maps  
  \item Employing a 3D U-Net to capture spatio-temporal consistency  
  \item Incorporating multi-modal conditions to improve prediction accuracy and quality  
\end{itemize}

In this work, we propose an inpainting method for satellite-based precipitation maps based on diffusion models --- a rapidly advancing class of generative models known for stable training and high-quality generation. Unlike conventional approaches, our method employs a three-dimensional U-Net architecture with 3D convolutions to simultaneously capture the spatial patterns and temporal dynamics of precipitation fields. For training, we utilize the ERA5 hourly precipitation dataset, which provides complete precipitation maps without missing regions, and simulate unobserved areas by applying satellite observation masks derived from GSMaP.

Recently, diffusion models have been applied in the geoscience field \cite{li2024generative, ling2024diffusion, mardani2025residual, allen2025end}. Li et al. reported that diffusion models can serve as an alternative to current ensemble forecasting methods, achieving high prediction accuracy with lower computational costs \cite{li2024generative}. Allen et al. applied diffusion models to global weather prediction in an end-to-end framework, demonstrating performance comparable to existing methods that have been carefully designed and tuned by human experts \cite{allen2025end}. Ling et al. showed that diffusion models can downscale 180 years of surface variables in East Asia and effectively evaluate uncertainty through stochastic generation, which conventional deterministic methods cannot achieve \cite{ling2024diffusion}. Mardani et al. also employed diffusion models to predict residuals and refine deterministic first guesses, enabling the downscaling of several meteorological variables \cite{mardani2025residual}. Thus, diffusion models represent a promising approach in the geosciences due to their strong expressive capacity and highly stable training.

Therefore, we adopt a diffusion model for precipitation mapping and formulate the inpainting task within a conditional diffusion framework. Specifically, to guide the inpainting process, our method leverages multiple conditional inputs: (1) global infrared cloud satellite images, (2) topographic data, (3) latitude-longitude grids, and (4) physical time information. 
We use a 3D convolutional condition encoder to transform these variables into multi-scale features, and inject them into each U-Net encoder block.
As we will later describe in the sensitivity analysis results, these conditions strongly influence the inpainting process, enhancing both accuracy and quality.

\subsection{Paper Structure}
The remainder of the paper is organized as follows. Section 2 describes the data and conditional information used in this study. Section 3 details the proposed diffusion model, including the Denoising Diffusion Probabilistic Model (DDPM), its extension to conditional generation, the 3D U-Net architecture, and the conditioning data. Section 4 presents the experimental results, including the sensitivity analysis and GSMaP inpainting experiments. Finally, Section 5 concludes the paper and outlines directions for future research.

\section{Method}
\label{sec:method}
\subsection{Diffusion Models}
Diffusion models have attracted attention since the early 2020s as a powerful class of generative models. Generative models generate data by learning a mapping from latent variables --- sampled from a tractable distribution (e.g., Gaussian) --- to the target data distribution. 

Prior to the rise of diffusion models, the primary generative models were Variational Autoencoders (VAEs) \cite{kingma2013auto} and Generative Adversarial Networks (GANs) \cite{goodfellow2014generative}. VAEs are known for their stable training, achieved by simply maximizing the evidence lower bound (ELBO), but they often produce blurred outputs. In contrast, GANs can generate high-quality images when training succeeds, yet they are notorious for instability and mode collapse. Diffusion models, by comparison, offer both stable training and consistent generation of high-quality samples.

A prominent variant of diffusion models is the DDPM \cite{ho2020denoising}, which consists of two sequential processes: a forward diffusion process and a reverse diffusion process. In the forward process, Gaussian noise is gradually added to an original data sample \( x_{0} \) until it is completely corrupted at terminal time \( T \). This process is defined by
\begin{align}
    \label{fd_step}
    q(x_{t}\mid x_{t-1}) 
    &= \mathcal{N}\bigl(x_{t}; \sqrt{1-\beta_{t}}x_{t-1}, \beta_{t}I\bigr) \nonumber \\
    \;\Longleftrightarrow\;
    x_{t} &= \sqrt{1-\beta_{t}}x_{t-1} + \sqrt{\beta_{t}}\epsilon_{t},
    \quad \epsilon_{t}\sim\mathcal{N}(0,I), \\[1ex]
    \label{fd_cum}
    q(x_{t}\mid x_{0})   
    &= \mathcal{N}\bigl(x_{t}; \sqrt{\bar{\alpha}_{t}}x_{0}, (1-\bar{\alpha}_{t})I\bigr) \nonumber \\
    \;\Longleftrightarrow\;
    x_{t} &= \sqrt{\bar{\alpha}_{t}}x_{0} + \sqrt{1-\bar{\alpha}_{t}}\epsilon,
    \quad \epsilon\sim\mathcal{N}(0,I).
\end{align}
where \(\alpha_{t}=1-\beta_{t}\), \(\bar{\alpha}_{t}=\prod_{s=1}^{t}\alpha_{s}\), and \(\mathcal{N}(\mu,\sigma^2 I)\) denotes an isotropic Gaussian distribution with mean \(\mu\) and variance \(\sigma^2 I\).

In the reverse process, the model begins with the pure noise sample \( x_{T} \) and iteratively removes noise to recover \( x_{0} \). This process is parameterized by a network \(\epsilon_{\theta}(x_{t},t)\) that predicts the noise at time \( t \) from noisy sample \(x_{t}\). When the reverse diffusion is defined by
\begin{equation}
    \label{reverse_diff}
    p_{\theta}(x_{t-1}\mid x_{t})
    = \mathcal{N}\bigl(x_{t-1}; \mu_{\theta}(x_{t},t), \sigma_t^2 I\bigr),
\end{equation}
then the mean of the reverse transition is modeled as
\[
\mu_{\theta}(x_{t},t)
= \frac{1}{\sqrt{1-\beta_{t}}}\Bigl(x_{t}
- \frac{\beta_{t}}{\sqrt{1-\bar{\alpha}_{t}}}\,\epsilon_{\theta}(x_{t},t)\Bigr)
\]
and we set \(\sigma_{t}=\sqrt{\beta_{t}}\). Suppose we denote the true data distribution by \(p_{\text{data}}\). Then the network \(\epsilon_{\theta}\) is trained by
\begin{equation}
    \label{diffusion_loss}
    L_{\epsilon}(\theta)
= \mathbb{E}_{x_{0}\sim p_{\text{data}}, \epsilon\sim\mathcal{N}(0,I), t\sim\mathcal{U}(1,T)}
  \Bigl\lVert \epsilon
  - \epsilon_{\theta}\bigl(\sqrt{\bar{\alpha}_{t}}x_{0}
  + \sqrt{1-\bar{\alpha}_{t}}\epsilon, t\bigr)\Bigr\rVert_{2}^{2},
\end{equation}
where \(\epsilon\sim\mathcal{N}(0,I)\).

For sampling, one draws \( x_{T}\sim\mathcal{N}(0,I)\) and then for \(t=T,\dots,1\),
\begin{equation}
    \label{diff_sampling}
    x_{t-1}
= \frac{1}{\sqrt{\alpha_{t}}}\Bigl(x_{t}
  - \frac{1-\alpha_{t}}{\sqrt{1-\bar{\alpha}_{t}}}\epsilon_{\theta}(x_{t},t)\Bigr)
  + \sqrt{\beta_{t}}z,
\end{equation}
where \(z\sim\mathcal{N}(0,I)\) for \(t>1\) and \(z=0\) for \(t=1\).

We can convert this noise prediction process (\(\epsilon\)-prediction) to velocity prediction (\(v\)-prediction) \cite{salimans2022progressive}. Define
\begin{equation}
    \label{v_def}
    v_{t}
    = \sqrt{\bar{\alpha}_{t}}\,\epsilon
      - \sqrt{1-\bar{\alpha}_{t}}\,x_{0},
    \quad
    \epsilon\sim\mathcal{N}(0,I),
\end{equation}
and train a network \(v_{\theta}(x_{t},t)\) via
\begin{equation}
    \label{v_loss}
    L_{v}(\theta)
= \mathbb{E}_{x_{0}\sim p_{\text{data}}, \epsilon\sim\mathcal{N}(0,I), t\sim\mathcal{U}(1,T)}
  \Bigl\lVert v_{t}
  - v_{\theta}\bigl(\sqrt{\bar{\alpha}_{t}}x_{0}
  + \sqrt{1-\bar{\alpha}_{t}}\epsilon, t\bigr)\Bigr\rVert_{2}^{2}.
\end{equation}
From equation~\eqref{v_def}, one can recover
\begin{equation}
\begin{aligned}
    \hat{\epsilon}_{t}
    &= \sqrt{\bar{\alpha}_{t}}v_{\theta}(x_{t},t)
       + \sqrt{1-\bar{\alpha}_{t}}x_{t},\\
  \hat{x}_{0}
    &= \frac{x_{t}-\sqrt{1-\bar{\alpha}_{t}}\hat{\epsilon}_{t}}
           {\sqrt{\bar{\alpha}_{t}}}.
\end{aligned}
\label{reconstructions}
\end{equation}
and substitute \(\hat\epsilon\) into equation~\eqref{diff_sampling}. Empirically, \(v\)-prediction often yields higher-fidelity samples and more stable scaling across noise levels.

Sampling can be derived from equation~\eqref{diff_sampling}. Starting from \(x_{T}\sim\mathcal{N}(0,I)\) and iterating for
\(t=T,\ldots,1\), we substitute the reconstruction of
\(\hat\epsilon\) in equation~\eqref{reconstructions} into the
\(\epsilon\)-based sampling step \eqref{diff_sampling}.  This yields
\begin{equation}
\label{v_sampling}
x_{t-1}
  = \frac{1}{\sqrt{\alpha_{t}}}\!
    \left(
      x_{t}
      - \frac{1-\alpha_{t}}{\sqrt{1-\bar{\alpha}_{t}}}
        \hat{\epsilon}_t
    \right)
    + \sqrt{\beta_{t}}z,
\end{equation}
where \(z\sim\mathcal{N}(0,I)\) for \(t>1\) and \(z=0\) for
\(t=1\). 

The framework described above corresponds to \emph{unconditional generation}. In contrast, \emph{conditional generation} --- for instance, in the context of inpainting --- extends the noise prediction network to
\[
  v_{\theta}(x_{t}, c, t),
\]
by incorporating the noisy sample \(x_{t}\), the timestep \(t\), and the condition \(c\). In this setting, the model no longer learns the unconditional distribution \(p_{\text{data}}(x)\) but instead approximates the conditional distribution \(p_{\text{data}}(x\mid c)\) by minimizing a divergence (in DDPM, the Kullback--Leibler divergence) between the true conditional data distribution and the model distribution. This probabilistic formulation naturally captures the stochastic uncertainty in precipitation maps and, by explicitly modeling this uncertainty, we can improve predictive performance which typically outperforms deterministic regression from~$c$ to~$x$.

\begin{figure}[tb]
  \centering
  \includegraphics[width=0.8\textwidth]{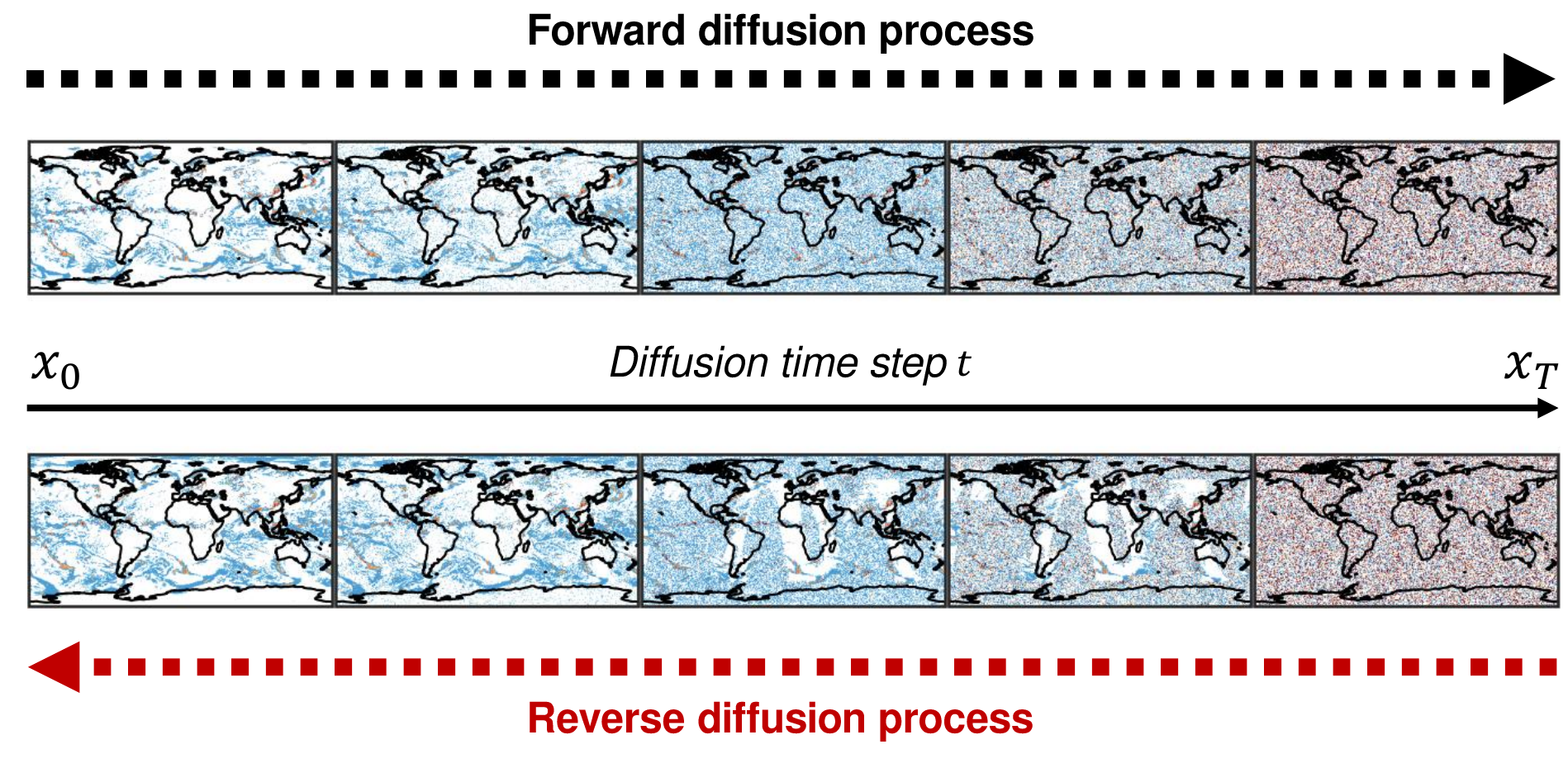}
  \caption{An example of forward and reverse diffusion processes applied to a global precipitation map. Forward diffusion corrupts the precipitation map sequence \(x_{0}\) by successive noise additions (upper), and reverse diffusion reconstructs the predicted sequence \(\hat{x}_{0}\) through iterative denoising, restoring observed regions to their original values as detailed in Algorithm 4 of the supporting information (lower).}
  \label{fig:diffusion_process}
\end{figure}

Figure~\ref{fig:diffusion_process} illustrates an example of the forward and reverse diffusion processes applied to a global precipitation map.. The upper side depicts the progressive corruption of the precipitation map sequence \(x_{0}\) through the addition of noise, while the lower side shows the reconstruction of the predicted sequence \(\hat{x}_{0}\) via iterative denoising.

\subsection{Latitude weight}
Global precipitation maps are rendered in the Plate Carr\'ee projection, which over-represents areas near the poles relative to those near the equator. To account for this distortion, we incorporate a latitude-weighting factor into the loss function, following the formulation introduced in ClimaX \cite{nguyen2023climax}:
\begin{equation}
  \label{eq:lat_weight}
     w(\phi_h) = \varepsilon + (1 - \varepsilon)\frac{\cos(\phi_h)}{\displaystyle \mathbb{E}_{\phi}\!\bigl[\cos(\phi)\bigr]},
    \quad
    \mathbb{E}_{\phi}[\cos(\phi)] = \frac{1}{H}\sum_{h=1}^{H}\cos(\phi_h),
\end{equation}
where \(\phi_{h}\) denotes the latitude at each grid row \(h \), \(H\) is the total number of latitude bands, and \(\varepsilon\) is a small constant that prevents completely discarding polar contributions. This weighting ensures that errors at mid and low latitudes are not fully diminished by the projection's polar stretching.  

\subsection{Training Procedure}
To simulate the presence of unobserved regions in GSMaP, pseudo-GSMaP data are generated as follows. Both the masked precipitation data and the condition inputs are normalized to the range \([0,1]\) in valid regions, and invalid (masked) regions are assigned a value of \(-1\). We use this \(-1\) value as the fundamental missing observation value throughout our study. The masking algorithm is described in Algorithm 2 of the supporting information. 

Now we can train the precipitation map inpainting DDPM using $v$-prediction. By using equation~\eqref{v_loss}, and on-the-fly generation of masked precipitation map via Algorithm 2 of the supporting information, we can train $v$-prediction DDPM.

\subsection{Network Architecture}
\begin{figure}[tb]
  \centering
  \includegraphics[width=1.0\textwidth]{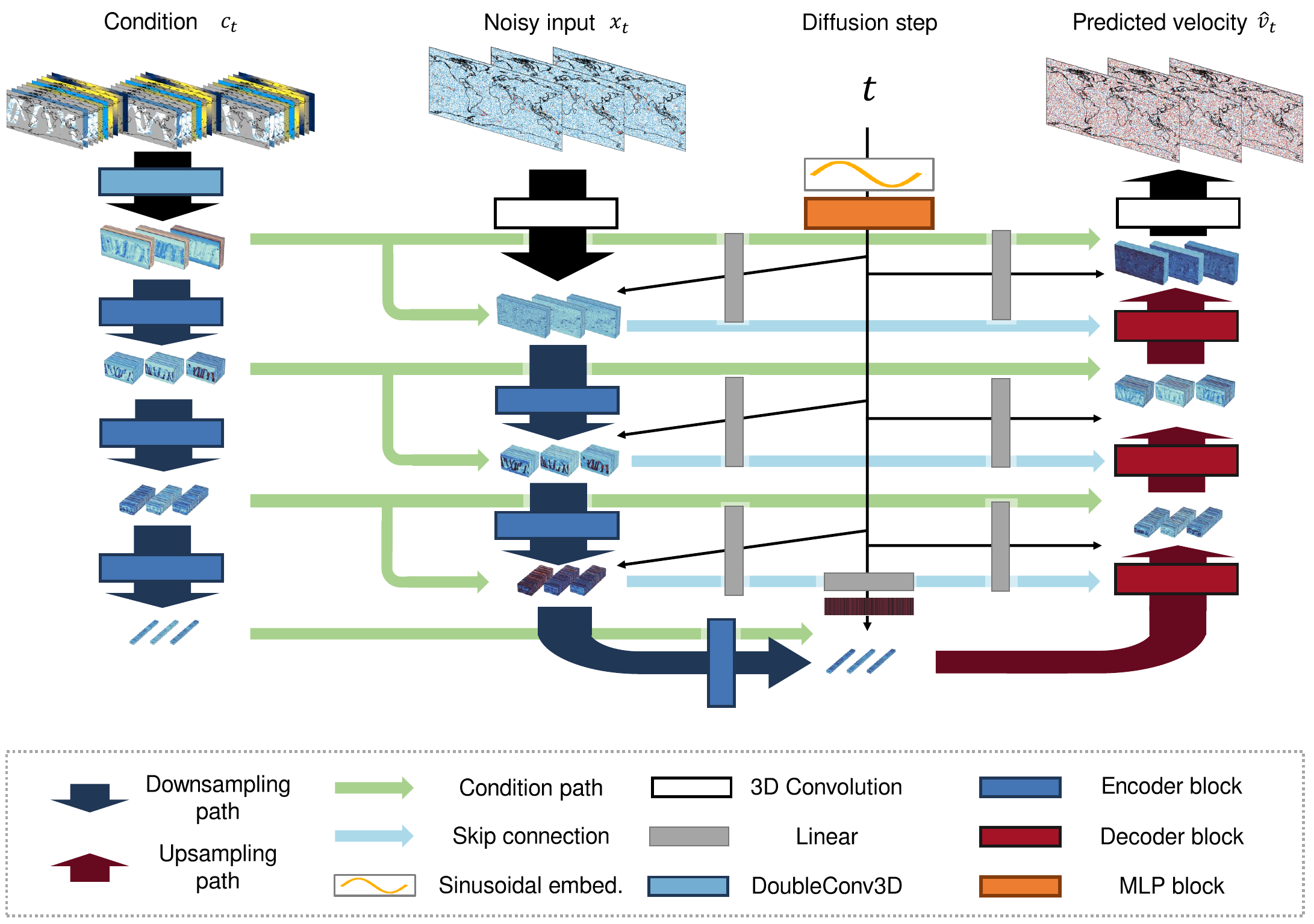}
  \caption{The 3D U-Net architecture in our study.}
  \label{fig:unet}
\end{figure}

We adopt a 3D U-Net~\cite{cciccek20163d}, which is a volumetric extension of the 2D U-Net~\cite{ronneberger2015u}, as shown in Figure~\ref{fig:unet}.
The skip connections in U-Net preserve fine-grained detail throughout the reverse diffusion process by directly passing encoder features to the decoder; this property has established U-Net as the de-facto standard backbone for diffusion models~\cite{ho2020denoising, nichol2021improved}.

While typical image diffusion models, such as Stable Diffusion~\cite{rombach2022high}, employ a 2D U-Net, our task requires 3D recognition capability to inpaint precipitation map sequences with spatio-temporal consistency. Schmutz et al. demonstrated the effectiveness of the 3D U-Net on historical weather data, reporting lower reconstruction errors compared to 2D baselines, which suggests that 3D recognition is also important in meteorological applications \cite{schmutz2024enhanced}. Taken together, these results indicate that a 3D U-Net is essential for accurate precipitation sequence inpainting in DDPM.

The U-Net architecture consists of three parts: its encoder, its decoder, and the condition encoder. The encoder takes the noisy input \(x_{t}\) and compresses it into a bottleneck representation via a single convolutional layer followed by three encoder blocks. The decoder then produces the velocity prediction \(\hat{v}_{t}\) by processing the bottleneck representation together with the encoder feature maps --- transferred via skip connections --- using three decoder blocks and a final convolutional layer. 

The diffusion time step \(t\) is first transformed through a sinusoidal embedding and a multi-layer perceptron (MLP) block; the resulting time embedding is incorporated into each stage of the encoder and decoder through linear projections. This time condition shifts the encoder and decoder features to control the U-Net output at the specified \(t\). 
The condition encoder extracts reference information through one DoubleConv3D block followed by three encoder blocks. 

The encoded features are added to the feature maps of both the main encoder and decoder to guide the prediction to be conditioned on \(\bar{c}\).
In our method, this architecture realizes the function
\(
  v_{\theta}(x_{t}, \bar{c}, t)
\)
as defined in Algorithms 3 and 4 (see the supporting information).
 Implementation details of each block are provided in the supporting information.

\subsection{Condition Data}
The condition input comprises ten channels: the masked precipitation map, a binary mask, two infrared (IR) brightness-temperature bands used in GSMaP, physical time embeddings, topographic elevation, and four latitude-longitude coordinate maps.  Given a prediction sequence of length $L$ and spatial resolution $(H,W)$, the condition tensor has shape $(10, L, H, W)$.  All valid data are scaled to $[0,1]$, whereas missing or masked values are represented by $-1$.  Both the IR dataset and topographic elevation data are preprocessed offline via the logistic transform \eqref{eq:sigmoid}-\eqref{eq:Adef} and stored in $[0,1]$; therefore, no additional scaling is required at runtime.

\subsubsection{Dynamic Condition Data}
Three types of dynamic inputs vary over time steps $n=0,\dots,L-1$:

\paragraph{Masked Precipitation Map}
Let $x_0^n$ denote the true precipitation at time step $n$ and $m_n\in\{0,1\}$ the corresponding observation mask (Algorithm 2 of the supporting information).  We define
\[
\bar x_0^n = m_n\,x_0^n + (1 - m_n)\,(-1),
\]
so that observed values remain within $[0,1]$ and all masked pixels are set to $-1$.  The sequence $\{\bar x_0^n\}_{n=0}^{L-1}$ constitutes the first channel of the condition tensor.

\paragraph{Binary Mask}
We encode the mask as
\[
M_{n,h,w} =
\begin{cases}
+1, & \text{newly masked region (to inpaint)},\\
0,  & \text{observed region},
\end{cases}
\]
for each time index $n$ and spatial location $(h,w)$.  This channel enables the model to distinguish between known and missing pixels.

Because precipitation fundamentally occurs where clouds are present, this condition is expected to improve the accuracy of precipitation inpainting in regions lacking valid precipitation observations. Valid infrared brightness temperature values are mapped into $[0,1]$ via the logistic function
\begin{equation}
\label{eq:sigmoid}
S(x) =
\begin{cases}
\dfrac{1}{1 + \exp\bigl(-A\,(x - \tfrac{x^h + x^l}{2})\bigr)}, 
& x\ \text{is valid},\\[4pt]
-1, & x\ \text{is missing or invalid},
\end{cases}
\end{equation}
where the steepness parameter $A$ is chosen such that $S(x^h)=s_h$ and $S(x^l)=s_l$:
\begin{equation}
\label{eq:Adef}
A = \frac{\ln\!\bigl(\tfrac{s_{h}}{1 - s_{h}}\bigr)\;-\;\ln\!\bigl(\tfrac{s_{l}}{1 - s_{l}}\bigr)}{x^{h} - x^{l}},
\end{equation}
with thresholds
\[
x^{l} = 270\ \mathrm{K},\quad
x^{h} = 230\ \mathrm{K},\quad
s_{l} = 0.2,\quad
s_{h} = 0.8.
\]

\paragraph{Physical Time Embedding}
To capture multi-scale temporal cycles, each physical time $\tau_n$ (in seconds) is embedded as follows.  First, compute fractional days since a reference time $\tau_0$:
\[
d_n = \frac{\tau_n - \tau_0}{86400}\quad[\text{days}].
\]
For five cycles with different wavelengths of $C_j\in\{7,\,30,\,365,\,3650,\,36500\}$ days, define
\[
\sin_{n,j} = \sin\!\Bigl(2\pi\frac{d_n}{C_j}\Bigr),\quad
\cos_{n,j} = \cos\!\Bigl(2\pi\frac{d_n}{C_j}\Bigr),
\]
and subsequently scale into $[0,1]$ by
\[
\sin^{\prime}_{n,j} = \frac{\sin_{n,j} + 1}{2},\quad
\cos^{\prime}_{n,j} = \frac{\cos_{n,j} + 1}{2}.
\]
Concatenation yields
\[
\mathbf{v}_n = \bigl[\sin^{\prime}_{n,1},\, \cos^{\prime}_{n,1},\dots,\sin^{\prime}_{n,5},\, \cos^{\prime}_{n,5}\bigr]\in[0,1]^{10}.
\]
Each vector \(\mathbf{v}_n\) is expanded to the spatial grid by repeating it \(\lceil H/10\rceil\) times along the latitude axis and truncating to length \(H\), then replicating each of the \(H\) rows across the \(W\) longitude columns to form an \((H\times W)\) map.

\subsubsection{Static Condition Data}
Static inputs are repeated $L$ times to align with the sequence length. Each condition is described as follows.

\paragraph{Topography}
We use topographical data, preprocessed into $[0,1]$ via \eqref{eq:sigmoid}-\eqref{eq:Adef} with
\[
x^{l} = 200\ \mathrm{m},\quad
x^{h} = 2000\ \mathrm{m},\quad
s_{l} = 0.2,\quad
s_{h} = 0.8,
\]
as one of the condition channels. We expect this condition to improve prediction accuracy in regions of orographic precipitation.

\paragraph{Latitude-Longitude Grids}
Define a regular grid
\[
\varphi_h = -\frac{\pi}{2} + \frac{h}{H-1}\,\pi,\quad
\lambda_w = -\pi + \frac{w}{W-1}\,2\pi,
\]
for $h=0,\dots,H-1$ and $w=0,\dots,W-1$, covering $90^\circ\mathrm{S}$-$90^\circ\mathrm{N}$ in latitude and $180^\circ\mathrm{W}$-$180^\circ\mathrm{E}$ in longitude.  We compute four normalized static maps:
\[
C^{\varphi}_{h,w} = \frac{\cos\varphi_h + 1}{2},\quad
S^{\varphi}_{h,w} = \frac{\sin\varphi_h + 1}{2},\quad
S^{\lambda}_{h,w} = \frac{\sin\lambda_w + 1}{2},\quad
C^{\lambda}_{h,w} = \frac{\cos\lambda_w + 1}{2},
\]
each broadcast to shape $(L,1,H,W)$. We expect this condition to function analogously to the positional encoding in Transformers \cite{vaswani2017attention}, enabling the model to recognize the spatial position to which each generated pixel corresponds.

\subsection{Computational Environment}
All experiments were performed on a single workstation configured as follows:
\begin{itemize}
    \item \textbf{CPU:} Intel Core i7-12700K (12 cores and 20 threads) $\times$ 1
    \item \textbf{GPU:} NVIDIA RTX A6000 (48\,GB VRAM) $\times$ 1
    \item \textbf{Memory:} 128\,GB
\end{itemize}

\section{Experimental settings}
We employed the following datasets in this study. For training, we used hourly data from 00:00 UTC on 1 January 2020 to 23:00 UTC on 31 December 2023; for evaluation, we considered the entire calendar year of 2024 (details on the selected hours are provided below). All data were regridded to a resolution of $(H,W)=(180,360)$, where $H$ and $W$ denote the latitude and longitude dimensions, respectively, spanning from $90^\circ\mathrm{S}$ to $90^\circ\mathrm{N}$ and from $180^\circ\mathrm{W}$ to $180^\circ\mathrm{E}$. We used $L=3$ as the prediction length in the experiment. We set $\varepsilon = 0.01$ for the latitude weight in equation~\eqref{eq:lat_weight}.
During training, one channel in the condition is randomly omitted by filling it with \(-1\) --- which denotes an invalid region in our condition setting --- with probability \(p_{\rm cond}\), in order to prevent the model from becoming overly dependent on any specific channel.

\subsection{Precipitation map dataset}
We used the ERA5 single-level hourly precipitation dataset \cite{era5precip} to obtain complete global precipitation fields. To compress precipitation values into the unit interval \([0,1]\), we applied an exponential transform. First, given a reference precipitation value \(x_{p}\) and a saturation probability \(p_{s}\), we compute the scale parameter
\[
  k \;=\;\frac{x_{p}}{-\ln\bigl(1 - p_{s}\bigr)}.
\]
The forward transform is defined as
\begin{equation}
  \label{eq:exp_transform}
  y \;=\; T(x)\;=\;1 - \exp\!\Bigl(-\frac{x}{k}\Bigr),
\end{equation}
and the inverse transform is
\begin{equation}
  \label{eq:inv_exp_transform}
  x \;=\; T^{-1}(y)\;=\;-\,k\,\ln\bigl(1 - y\bigr).
\end{equation}
  
We employed these transforms in our experiments to preprocess the dataset and to reconstruct predictions in the original physical units from the normalized model outputs.
In our experiments, we set the reference precipitation value to \(x_{p} = 5.0\ \mathrm{mm\,h^{-1}}\) and the saturation probability to \(p_{s} = 0.99\).  
To convert the unit of ERA5 precipitation from $\mathrm{m\,h^{-1}}$ to $\mathrm{mm\,h^{-1}}$, we multiplied the values by $1000$.
To simulate missing regions, we applied the GSMaP satellite precipitation product \cite{kubota2020global} as described below.

\subsection{Mask generation}
Following the GSMaP data specification \cite{gsmap_manual}, we generated a binary mask of \emph{invalid} pixels according to:
\begin{itemize}
  \item \emph{No TRMM/TMI observation}: \texttt{satelliteInfoFlag}~$\leq 1$.
  \item \emph{Missing observation}: \texttt{hourlyPrecipRate}~$<0$.
\end{itemize}
The mask is defined as the union of these two sets (i.e.,\ pixels satisfying at least one of the above conditions). Hereafter, “GSMaP precipitation” denotes the precipitation field with all masked (invalid) pixels removed.

\subsection{Infrared cloud imagery}
We incorporated two infrared brightness-temperature channels from the GPM Merged IR product \cite{nasa_ir}, covering latitudes between $60^\circ\mathrm{S}$ and $60^\circ\mathrm{N}$.

\subsection{Topography}
Bedrock elevation data were obtained from the ETOPO dataset \cite{etopo2022}, at a spatial resolution of 30 arc-seconds.

\subsection{Evaluation metrics}
\label{data_selection}
To reduce computational cost, we sampled twelve three-hour windows --- one per month --- rather than process every possible interval.  All windows are half-open intervals \([t_{\rm start},t_{\rm end})\), so the end time itself is excluded.  Specifically, we evaluated:
\[
  (00{:}00\text{--}03{:}00,\;01/01),\quad
  (02{:}00\text{--}05{:}00,\;02/01),\;\dots,\;
  (22{:}00\text{--}01{:}00,\;12/01\text{--}12/02).
\]
For each of these twelve windows, generative models produced \(K=16\) independent predictions from distinct latent seeds, and we used the ensemble mean as the deterministic prediction (deterministic baselines yield one output per window).  We then computed each metric (RMSE, MS-SSIM, BDI, TG-RMSE, CSI, eNLL, coverage, sharpness, CRPS) across these twelve samples and derived 95\% confidence intervals via bootstrap.

\section{Results}

\subsection{Selected Methods for Comparison}
To evaluate the effectiveness of our DDPM-based approach, we compared it against benchmark methods including both classical (non-deep) algorithms and modern deep learning-based approaches.

\subsubsection{Non-deep learning methods (Temporal Linear Interpolation).}
We adopt simple temporal linear interpolation (TLI) as our non-deep baseline. At each spatial location \((h,w)\), any missing pixel \(x_{t,h,w}\) is reconstructed by linearly interpolating between its two nearest observed neighbors in time.

Let
\[
\mathcal{O}_{h,w}
= \{\,t : x_{t,h,w}\text{ is observed}\}, 
\qquad
\mathcal{N}_{h,w}
= \{\,t : x_{t,h,w}\text{ is missing}\}.
\]
For each \(t\in\mathcal{N}_{h,w}\), define
\begin{align}
t_i &= \max\{\tau < t \mid \tau \in \mathcal{O}_{h,w}\},\\
t_j &= \min\{\tau > t \mid \tau \in \mathcal{O}_{h,w}\},\\
\hat{x}_{t,h,w}
    &= x_{t_i,h,w}
     + \frac{t - t_i}{t_j - t_i}
       \bigl(x_{t_j,h,w}-x_{t_i,h,w}\bigr).
\end{align}
If \(\lvert\mathcal{O}_{h,w}\rvert=1\), every missing frame is assigned that single observed value:
\[
\hat x_{t,h,w} = x_{t',h,w}, 
\quad t' = \text{the unique element of }\mathcal{O}_{h,w}.
\]
If \(\lvert\mathcal{O}_{h,w}\rvert=0\), the pixel remains missing. We implement TLI using NumPy's \texttt{interp} function.

To further fill spatial gaps, we introduce TLI-NS: after performing TLI, we apply OpenCV's Navier-Stokes inpainting algorithm \cite{bertalmio2001navier} to any remaining missing pixels.

\subsubsection{Deep learning methods}
\paragraph{Supervised U-Net.}  
We adapt our 3D U-Net architecture from the DDPM backbone for deterministic regression by removing all time-embedding modules. The network is trained to map a masked precipitation field directly to the corresponding full precipitation field. We refer to this method simply as ``U-Net.''

\paragraph{Variational Autoencoder.}  
We implement a conditional VAE \cite{kingma2013auto} by retaining the 3D U-Net as the decoder. A lightweight encoder compresses the masked input into the mean \(\mu\) and standard deviation \(\sigma\) of a Gaussian latent distribution \(\mathcal{N}(\mu,\sigma^2)\). During training, latent vectors are sampled via the reparameterization trick and upsampled to the precipitation map resolution before decoding; at inference, we draw from the Gaussian prior $N(0,I)$ and pass the latent vector along with the condition to the decoder.

\paragraph{Generative Adversarial Network.}  
We adopt the pix2pix framework \cite{isola2017image}, which augments a supervised U-Net generator with an adversarial loss. A PatchGAN discriminator guides the generator to refine high-frequency details, and the generator is optimized using a weighted sum of the L1 reconstruction and adversarial losses.

\paragraph{Diffusion-based models.}  
We evaluate standard DDPMs using both linear and cosine noise schedules. Sampling is performed with the original DDPM sampler, as well as the Denoising Diffusion Implicit Model (DDIM) sampler \cite{song2020denoising}. In addition, we include the Rectified Flow method \cite{liu2022flow}, employing a fourth-order Runge--Kutta solver during the sampling trajectory.

\subsection{Main Results}

\begin{figure}[tb]
  \centering
  \includegraphics[width=1.0\textwidth]{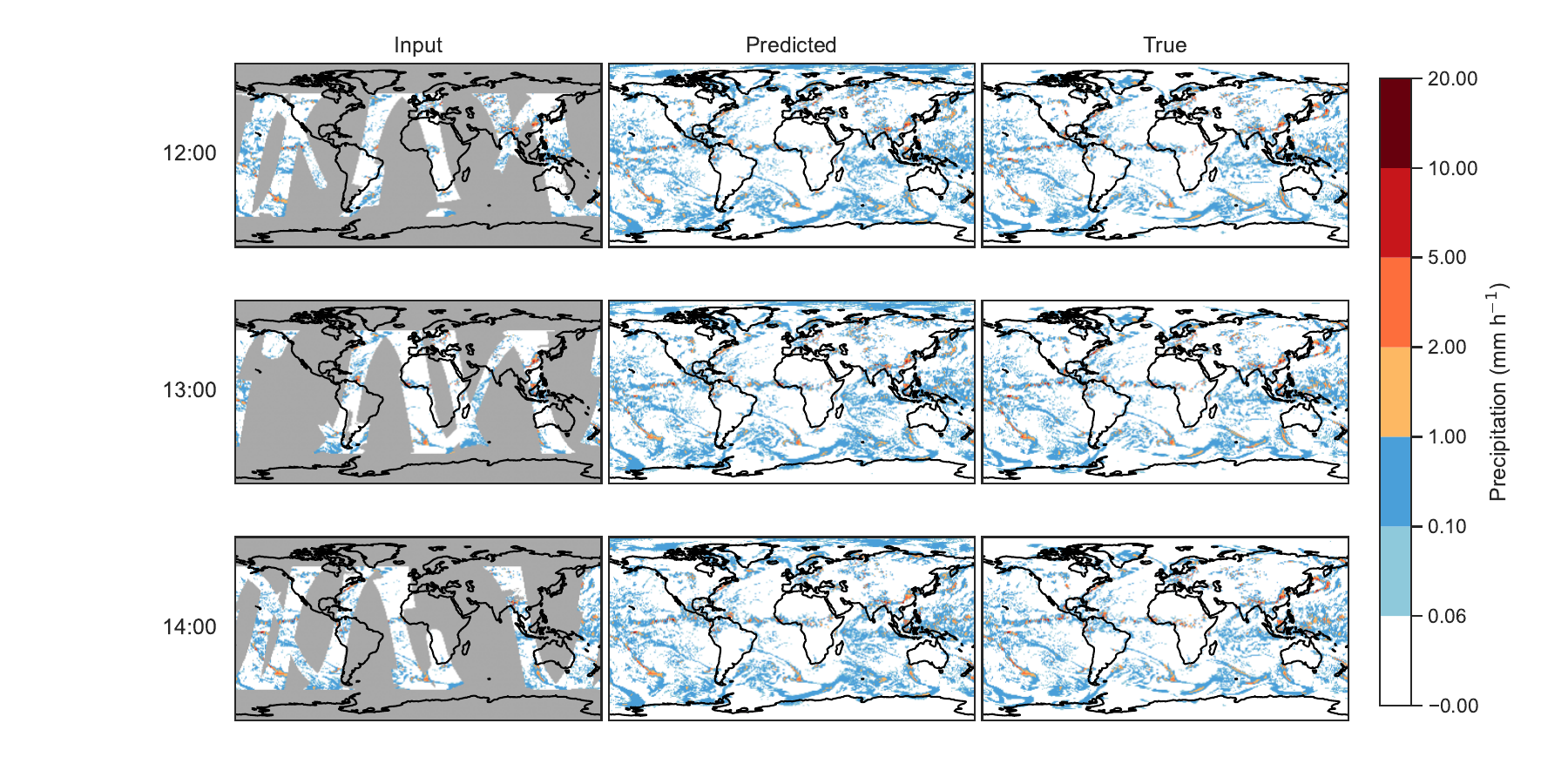}
  \caption{Comparison of the input masked sequence (left), the inpainted sequence (middle), and the ground truth sequence (right), using example data from the ERA5 precipitation maps between 12:00 and 14:00 UTC on July 1, 2024. Gray shading indicates missing regions.}
  \label{fig:comparison}
\end{figure}

We first evaluated the prediction performance using example data from the ERA5 precipitation maps between 12:00 and 14:00 UTC on July 1, 2024 (evaluation dataset). We employ a DDPM with a linear noise schedule and \(T=1000\). Figure~\ref{fig:comparison} compares the input masked sequence, the inpainted sequence, and the ground truth sequence. The results demonstrate that our method successfully inpaints the missing regions of the precipitation map sequence in a natural manner.

\begin{figure}[tb]
  \centering
  \begin{subfigure}[b]{0.48\textwidth}
    \centering
    \includegraphics[width=\textwidth]{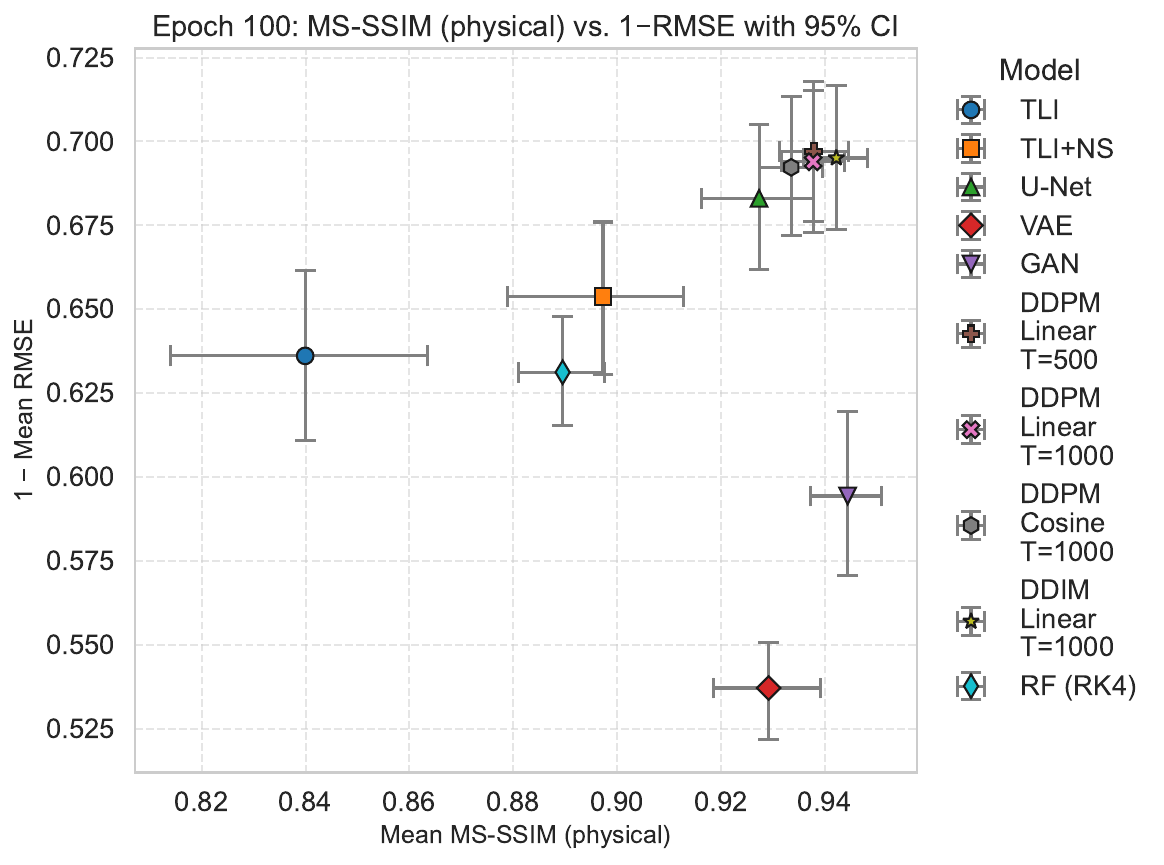}
    \caption{Image quality vs.\ prediction accuracy. The top right is best.}
    \label{fig:scatter_quality}
  \end{subfigure}
  \hfill
  \begin{subfigure}[b]{0.48\textwidth}
    \centering
    \includegraphics[width=\textwidth]{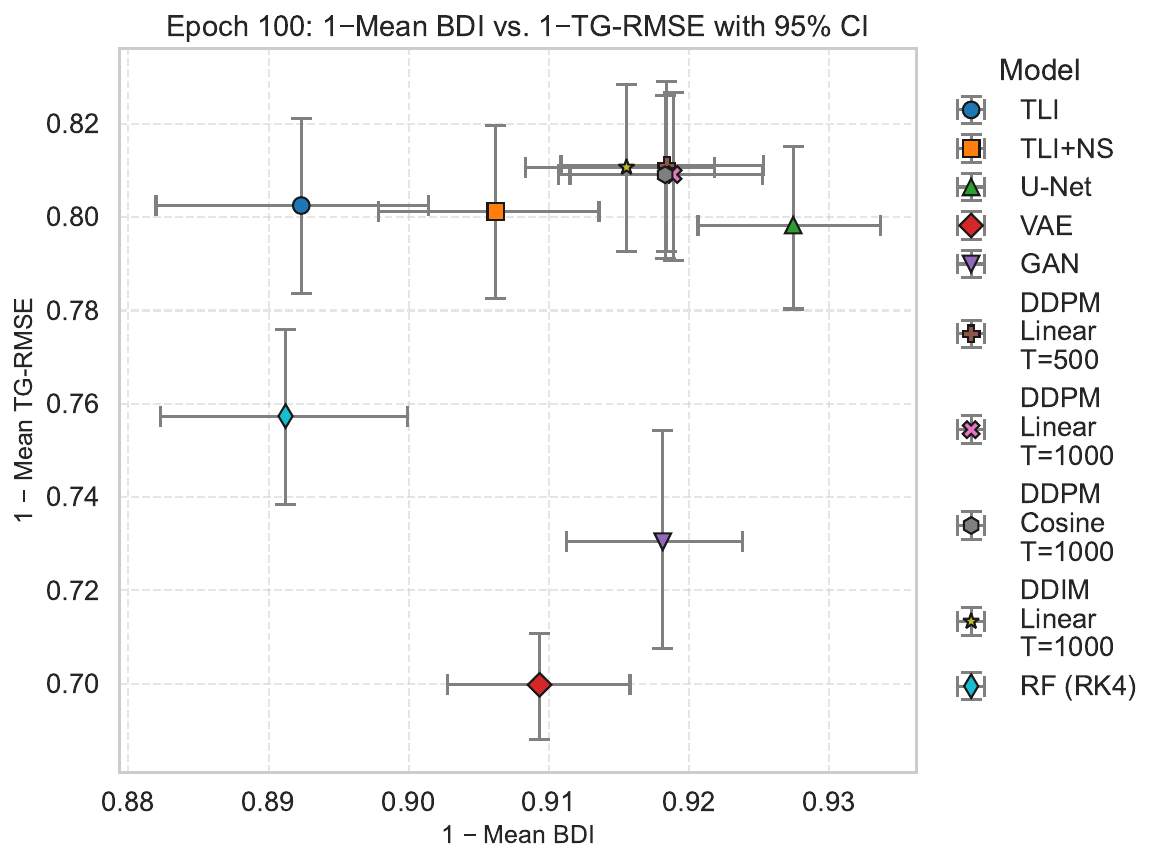}
    \caption{Spatial vs.\ temporal consistency. The top right is best and well-balanced.}
    \label{fig:scatter_consistency}
  \end{subfigure}
  \caption{Comparison of our method against alternative approaches along two evaluation axes, with vertical and horizontal error bars at each point denoting the 95\% confidence intervals.}

  \label{fig:scatter_combined}
\end{figure}

Next, we evaluated our method on the full evaluation dataset --- using the hour selection procedure described in Section~\ref{data_selection} --- and compared its performance with that of alternative approaches.
Figure~\ref{fig:scatter_quality} illustrates the scatter plot of mean MS-SSIM (x-axis) versus \(1 -\) mean RMSE (y-axis). Since RMSE is a lower-is-better metric, \(1 - \mathrm{RMSE}\) serves as a higher-is-better measure of prediction accuracy. MS-SSIM quantifies perceptual similarity to the ground truth, with higher values indicating better image quality. Consequently, the plot can be divided into four regions:
\begin{itemize}
  \item Lower left: low image quality and low prediction accuracy (worst case).
  \item Lower right: high image quality but low prediction accuracy (realistic but inaccurate outputs).
  \item Upper left: low image quality but high prediction accuracy (meaningful but visually degraded outputs).
  \item Upper right: high image quality and high prediction accuracy (best case).
\end{itemize}

As shown in Figure~\ref{fig:scatter_quality}, DDPMs (our method) achieve the highest image quality and prediction accuracy. Interestingly, the U-Net (conventional supervised learning) attains performance comparable to the DDPMs. In contrast, the VAE and GAN generate realistic sequences but suffer from low prediction accuracy. Rectified Flow fails to surpass the TLI+NS baseline. Overall, deep learning methods outperform non-deep baselines, with DDPMs emerging as the top-performing deep model. These results suggest that deep learning is a promising approach for precipitation map inpainting and that diffusion models attain state-of-the-art performance in weather-related tasks.

Figure~\ref{fig:scatter_consistency} presents the evaluation of spatial and temporal consistency. Since BDI is a lower-is-better metric, \(1 - \mathrm{BDI}\) serves as a higher-is-better measure of spatial consistency, quantifying the seamlessness between regions inside and outside the mask border. TG-RMSE assesses temporal smoothness of adjacent frames, representing temporal consistency. Since TG-RMSE is a lower-is-better metric, we define \(1 - \mathrm{TG\text{-}RMSE}\) as a higher-is-better measure of temporal consistency, quantifying the smoothness of adjacent frames.
Accordingly, the plot can be partitioned into four regions:
\begin{itemize}
  \item Lower left: low spatial and temporal consistency (worst case).
  \item Lower right: high spatial but low temporal consistency (spatial focus).
  \item Upper left: low spatial but high temporal consistency (temporal focus).
  \item Upper right: high spatial and temporal consistency (best case).
\end{itemize}

Inspection of Figure~\ref{fig:scatter_consistency} reveals that the methods divide into two groups: those prioritizing spatial consistency and those prioritizing temporal consistency. The former group, comprising U-Net, VAE, and GAN, relies primarily on spatial features (via 3D convolutions), resulting in higher spatial but lower temporal consistency. The latter group, including DDPM variants and the TLI-based method, emphasizes temporal over spatial consistency. While the inclusion of TLI in this group aligns with its design, the tendency of DDPMs to favor temporal consistency is particularly noteworthy. Both U-Net and DDPM achieve the best balance between the two dimensions.

\subsection{Sensitivity Analysis of Condition}
\renewcommand*{\arraystretch}{1.5}
\setlength{\extrarowheight}{4pt}

\begin{table}[tb]
  \centering
  \scriptsize
  \caption{Results of the sensitivity analysis conducted using the model at epoch 100 of our DDPM method with a linear noise schedule ($T=1000$). ``Full model'' refers to the original configuration (10 condition channels), while ``Minimum'' denotes the minimal-condition model trained from scratch. The remaining entries represent performance differences, $\Delta$, obtained by selectively omitting specific condition channels from the full model. In each cell, the upper value indicates the mean and the lower value indicates the 95\% confidence interval.}
  \label{tab:sensitivity}
  \begin{tabular}{@{}>{\raggedright\arraybackslash}m{2.5cm}>{\centering\arraybackslash}m{1.0cm}>{\centering\arraybackslash}m{1.8cm}>{\centering\arraybackslash}m{1.8cm}>{\centering\arraybackslash}m{1.8cm}>{\centering\arraybackslash}m{1.8cm}@{}}
    \toprule
    Condition & Contrib. & RMSE & MS-SSIM & TG-RMSE & BDI \\
    \midrule
    \textbf{Full model} &  & 0.306\newline $[0.286, 0.327]$ & 0.938\newline $[0.932, 0.943]$ & 0.191\newline $[0.173, 0.209]$ & 0.081\newline $[0.074, 0.088]$ \\
    \midrule
    \multicolumn{1}{r}{w/o masked precip} & 0.415 & 0.032\newline $[0.027, 0.037]$ & -0.025\newline $[-0.028, -0.022]$ & 0.020\newline $[0.015, 0.023]$ & 0.015\newline $[0.013, 0.017]$ \\
    \multicolumn{1}{r}{w/o mask} & 0.217 & 0.010\newline $[0.007, 0.014]$ & -0.012\newline $[-0.014, -0.010]$ & 0.006\newline $[0.005, 0.008]$ & 0.019\newline $[0.016, 0.023]$ \\
    \multicolumn{1}{r}{w/o IR} & 0.273 & 0.018\newline $[0.015, 0.021]$ & -0.035\newline $[-0.047, -0.027]$ & 0.003\newline $[0.001, 0.005]$ & 0.003\newline $[0.002, 0.005]$ \\
    \multicolumn{1}{r}{w/o time} & 0.004 & -0.001\newline $[-0.002, 0.001]$ & 0.000\newline $[-0.001, 0.001]$ & 0.000\newline $[-0.001, 0.001]$ & 0.002\newline $[0.000, 0.003]$ \\
    \multicolumn{1}{r}{w/o ETOPO} & 0.050 & 0.004\newline $[0.002, 0.007]$ & -0.004\newline $[-0.007, -0.002]$ & 0.001\newline $[-0.000, 0.003]$ & 0.001\newline $[-0.000, 0.003]$ \\
    \multicolumn{1}{r}{w/o latitude} & 0.028 & 0.002\newline $[0.000, 0.004]$ & -0.001\newline $[-0.003, -0.000]$ & 0.001\newline $[0.000, 0.003]$ & 0.001\newline $[-0.000, 0.003]$ \\
    \multicolumn{1}{r}{w/o longitude} & 0.013 & -0.000\newline $[-0.002, 0.001]$ & -0.002\newline $[-0.003, -0.001]$ & 0.000\newline $[-0.001, 0.001]$ & 0.001\newline $[0.000, 0.003]$ \\
    \midrule
    \multicolumn{1}{r}{w/o lat/lon} &  & 0.008\newline $[0.007, 0.009]$ & -0.005\newline $[-0.007, -0.003]$ & 0.003\newline $[0.002, 0.004]$ & 0.004\newline $[0.003, 0.005]$ \\
    \midrule
    \multicolumn{1}{r}{w/o dynamic conds} &  & 0.073\newline $[0.065, 0.083]$ & -0.087\newline $[-0.102, -0.074]$ & 0.027\newline $[0.023, 0.031]$ & 0.020\newline $[0.017, 0.023]$ \\
    \multicolumn{1}{r}{w/o static conds} &  & 0.017\newline $[0.014, 0.019]$ & -0.016\newline $[-0.020, -0.012]$ & 0.007\newline $[0.005, 0.009]$ & 0.007\newline $[0.006, 0.009]$ \\
    \midrule
    \multicolumn{1}{r}{Precip \& mask} &  & 0.049\newline $[0.044, 0.053]$ & -0.065\newline $[-0.073, -0.055]$ & 0.018\newline $[0.016, 0.020]$ & 0.016\newline $[0.014, 0.017]$ \\
    \midrule
    \textbf{Minimum} &  & 0.308\newline $[0.287, 0.328]$ & 0.939\newline $[0.931, 0.945]$ & 0.190\newline $[0.173, 0.208]$ & 0.081\newline $[0.074, 0.088]$ \\
    \bottomrule
  \end{tabular}
\end{table}

We performed a sensitivity analysis on the condition inputs of our trained model (DDPM with a linear noise schedule, $T=1000$). For each channel $i$ in the condition tensor, we replaced its values with $-1$ and measured the performance differences relative to the full model. If the condition input had multiple channels (e.g., the IR condition), all were replaced together. The contribution of each channel \( r_i \) was then quantified as follows:

\[
\begin{aligned}
M &= \{\mathrm{RMSE},\;\mathrm{MS\text{--}SSIM},\;\mathrm{TG\text{--}RMSE},\;\mathrm{BDI}\},\\[4pt]
s_m &=
\begin{cases}
+1, & m \in \{\mathrm{RMSE},\mathrm{TG\text{--}RMSE},\mathrm{BDI}\},\\[2pt]
-1, & m = \mathrm{MS\text{--}SSIM},
\end{cases}\\[6pt]
\mu_{\mathrm{full},m} &= \text{mean of metric }m\text{ for the full model},\\
\mu_{i,m} &= \text{mean of metric }m\text{ after removal of channel }i,\\[4pt]
\Delta_{i,m} &= s_m\bigl(\mu_{i,m}-\mu_{\mathrm{full},m}\bigr),\\[4pt]
\Delta_i &= \frac{1}{|M|}\sum_{m\in M}\Delta_{i,m},\\[4pt]
r_i &= \frac{\displaystyle\Delta_i}
                               {\displaystyle\sum_{j\in\mathcal{A}}\Delta_j},
\end{aligned}
\]
where $r_i$ denotes the normalized contribution of channel $i$. A positive $\Delta_i$ indicates performance degradation, whereas a negative $\Delta_i$ indicates improvement. The contribution rate $r_i$ reflects the relative influence of each condition, with larger values corresponding to greater impact.

Table~\ref{tab:sensitivity} summarizes the results. The key observations are:
\begin{itemize}
  \item ``Masked precip'': the masked precipitation sequence, which serves as the primary input for inpainting, is the most critical condition, indicating the model's heavy reliance on this information.
  \item ``Mask'': the binary mask identifying observed versus missing regions is the second most important condition, enabling the model to localize inpainting regions.
  \item ``IR'': infrared cloud imagery ranks third, providing essential cues for areas lacking valid pixels across all frames.
  \item ``Time'', ``ETOPO'', and ``Lat/Lon'': these temporal and static inputs exhibit negligible effects on performance.
  \item Simultaneous removal of latitude and longitude (``Lat/Lon'') does not degrade model accuracy.
\end{itemize}

In summary, the omission of temporal and static conditions (Time, ETOPO, Lat/Lon) does not impair prediction accuracy, whereas the removal of dynamic conditions (Masked precip, Mask, IR) leads to a substantial performance drop. Even retaining only the masked sequence and mask (``Precip \& mask'') results in noticeable degradation, highlighting the importance of IR information.

Based on these findings, we trained a reduced model conditioned solely on the masked sequence, mask, and IR channels, while keeping all other network architectures, training protocols, and evaluation settings identical. The results, presented at the bottom of Table~\ref{tab:sensitivity} under ``Minimum'', demonstrate that equivalent performance can be achieved using only these three condition inputs.  

\subsection{GSMaP inpainting using trained model on ERA}

\begin{figure}[tb]
  \centering
  \includegraphics[width=1.0\textwidth]{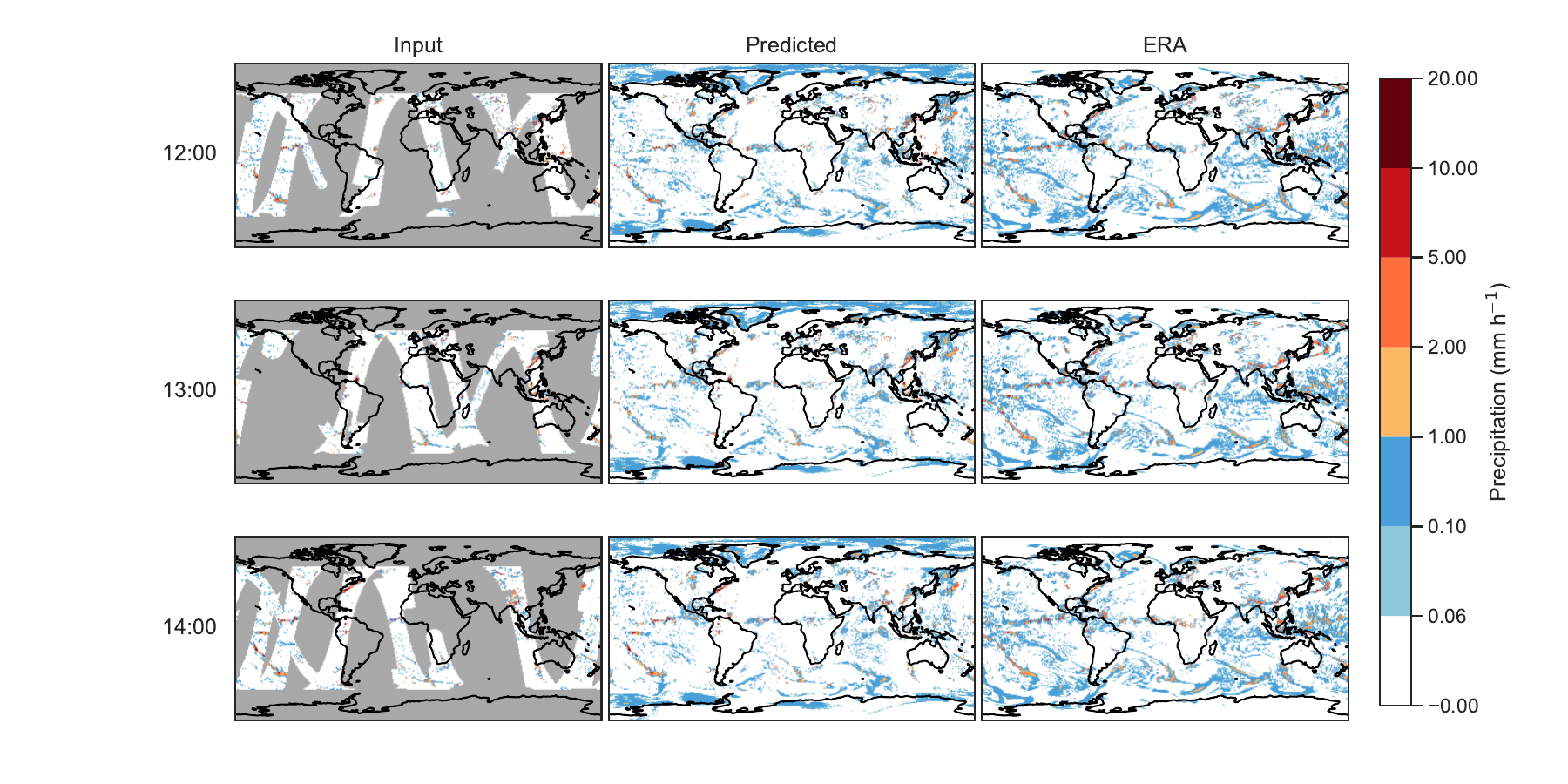}
  \caption{Comparison of the input GSMaP sequence (left), the inpainted sequence (middle), and the ERA sequence (right), using example data from the GSMaP precipitation maps between 12:00 and 14:00 UTC on July 1, 2024. Gray shading indicates missing regions.}
  \label{fig:comparison_gsmap}
\end{figure}

Finally, we applied our method --- using the DDPM model trained on ERA data with a linear noise schedule ($T=1000$) --- to real GSMaP sequences. The GSMaP data were transformed via the exponential transform in equation~\eqref{eq:exp_transform}, and the mask was extracted from the actual missing regions in the maps. Figure~\ref{fig:comparison_gsmap} shows the inpainting results: the left panel presents the original GSMaP sequence, the middle panel depicts the model's prediction, and the right panel displays the corresponding ERA sequence used as reference in the absence of a complete GSMaP ground truth. This setup constitutes a zero-shot inpainting scenario, confirming that our method can successfully reconstruct missing regions in GSMaP data.

\section{Conclusion}
High-fidelity precipitation mapping plays a pivotal role not only in advancing our meteorological understanding of the global water cycle but also in disaster mitigation and hydrological engineering design. However, global satellite-derived precipitation products --- such as GSMaP and IMERG --- often suffer from extensive missing data due to the orbital characteristics of polar-orbiting satellites that have microwave sensors. Conventional inpainting approaches primarily emphasize temporal consistency, which can introduce spatial artifacts. 

To overcome these limitations, we propose a conditional generative inpainting framework based on a denoising diffusion probabilistic model (DDPM) with a 3D U-Net backbone. Experimental results demonstrate that our method outperforms conventional statistical techniques and deep-learning baselines, including variational autoencoders (VAEs) and generative adversarial networks (GANs). 

Sensitivity analysis further reveals the critical importance of the binary observation mask and infrared (IR) cloud imagery, in addition to the masked precipitation sequence, for accurate reconstruction. Finally, we demonstrate that the model trained via our approach can effectively inpaint real GSMaP sequences. Future work will investigate more efficient network architectures and incorporate recent advances in generative modeling to further enhance inpainting performance.

\acknowledgments
This study was partially supported by the Japan Aerospace Exploration Agency (JAXA) Precipitation Measuring Mission (PMM), JSPS KAKENHI (grant nos. JP21H05002, JP25K17687 and JP25H00752), and the IAAR Research Support Program and VL Program of Chiba University. 

\section*{Conflict of Interest}
The authors declare no conflicts of interest relevant to this study.

\section*{Open Research}

\subsection*{Data Availability}

The ERA5 reanalysis data used for model training and evaluation were acquired from the Copernicus Climate Data Store (C3S; DOI: \url{10.24381/cds.adbb2d47}), with access dates ranging from 21 October 2024 to 12 June 2025. Hourly precipitation fields from the Global Satellite Mapping of Precipitation (GSMaP) project were obtained from the Japan Aerospace Exploration Agency (DOI: \url{10.57746/EO.01gs73bkt358gfpy92y2qns5e9}), accessed between 13 April 2025 and 1 May 2025. The GPM Merged IR data were retrieved via FTP from the Center for Environmental Remote Sensing at Chiba University (\url{ftp://gms.cr.chiba-u.ac.jp/pub/GPM_MergedIR/}), with access dates from 17 April 2025 to 19 April 2025. Topographic relief was taken from the ETOPO Global Relief Model \cite{etopo2022}, accessed 8 November 2024. No new observational data were generated in this study.

\subsection*{Software Availability}

All analysis and modeling code was implemented in Python 3.10.12 using PyTorch 2.6.0 and standard open-source libraries. To facilitate the peer-review process, the code is provided here as supporting information and will be deposited in a public repository upon the manuscript's publication.

\clearpage
\appendix

\section*{Supporting Information for ``Conditional Diffusion Models for Global Precipitation Map Inpainting''}

\noindent\textbf{Contents of this file}
\begin{enumerate}
  \item \textbf{Section 1.} Network implementation details
  \item \textbf{Section 2.} Diffusion-model hyper-parameters and noise schedules
  \item \textbf{Section 3.} Masking, DDPM training, and sampling algorithms
  \item \textbf{Section 4.} Baseline models and common training settings
  \item \textbf{Section 5.} Evaluation metrics used in the study
  \item \textbf{Figure S1.} Block diagrams of the 3D U-Net (encoder, decoder, MLP, DoubleConv3D + SE)
  \item \textbf{Table S1.} Hyper-parameter settings for the 3D U-Net
\end{enumerate}

\noindent\textbf{Introduction}
This supporting information provides all implementation details needed to reproduce our experiments.  
Section 1 describes the 3D U-Net backbone used across diffusion, VAE, GAN, and rectified flow baselines.  
Section 2 describes the linear and cosine noise schedules and other diffusion hyper-parameters.  
Section 3 presents pseudocode for data masking, DDPM training, and sampling.  
Section 4 details optimizer choices, data augmentation schemes, and the architectures of baseline models.  
Section 5 defines the evaluation metrics.  
Figure S1 visualises the core network blocks, while Table S1 lists every tunable hyper-parameter of the U-Net.  

\clearpage

\section{Network implementation details}
\label{ap:network}

\begin{figure}[tb]
  \centering
  \includegraphics[width=0.8\textwidth]{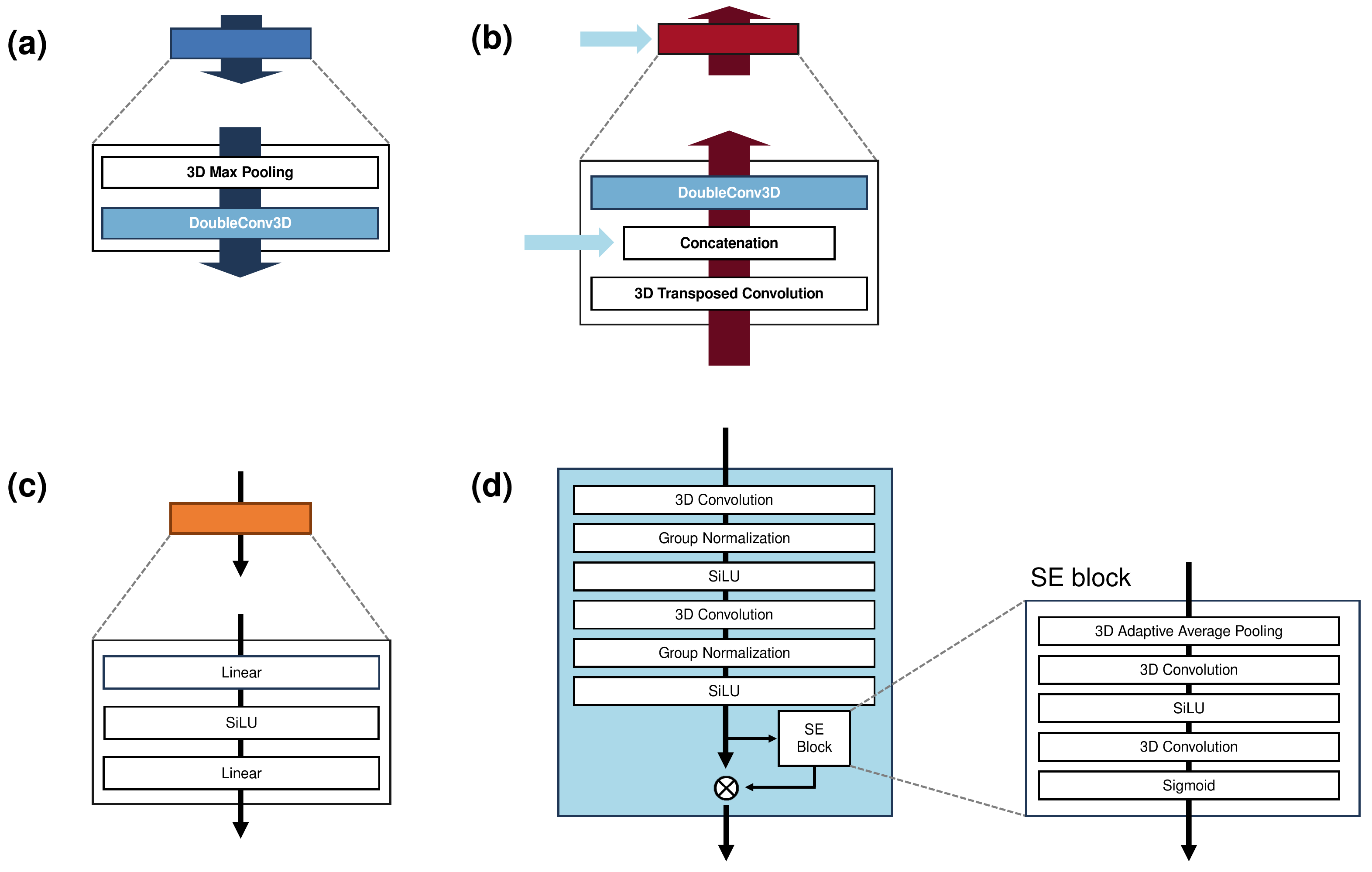}
  \caption{Details of each block. (a) Encoder block. (b) Decoder block. (c) MLP block. (d) DoubleConv3D and SE block.}
  \label{fig:block}
\end{figure}

We implement a three-dimensional U-Net for DDPM, illustrated in Figure 2 in main paper, comprising three core components: encoder blocks, decoder blocks, and an MLP block. The detailed layouts of these components are shown in Figure~\ref{fig:block}.

We adopt the notation \((B,C,T,H,W)\) to indicate tensor shapes --- batch size, channels, temporal length, height, and width, respectively. Hyper-parameter values are summarized in Table~\ref{tab:unet_hparams}.

\subsection{DoubleConv3D}
DoubleConv3D (Fig.~\ref{fig:block}d) consists of two sequential 3D convolutions, each followed by GroupNorm and SiLU activation. This design parallels the residual blocks in the original 2D U-Net \cite{ronneberger2015u}, substituting SiLU in place of ReLU \cite{elfwing2018sigmoid,ramachandran2017searching} for smoother gradient flow. A squeeze-and-excitation (SE) module \cite{hu2018squeeze} injects channel-wise recalibration: 3D adaptive average pooling aggregates spatio-temporal features into a C-dimensional descriptor, which is passed through a two-layer bottleneck MLP to produce per-channel modulation weights.

\subsection{Encoder block}
The encoder block (Fig.~\ref{fig:block}a) begins with 3D max pooling \((1,2,2)\), halving spatial dimensions: \((B,C,T,H,W)\to(B,C,T,H/2,W/2)\). A DoubleConv3D layer then doubles the channel count: \((B,C,T,H/2,W/2)\to(B,2C,T,H/2,W/2)\). Hence each encoder stage transforms \((B,C,T,H,W)\) into \((B,2C,T,H/2,W/2)\).

\subsection{Decoder block}
As shown in Figure~\ref{fig:block}b, each decoder block upsamples via a 3D transposed convolution \((2C\to C)\) with stride \((1,2,2)\), restoring spatial resolution: \((B,2C,T,H/2,W/2)\to(B,C,T,H,W)\). The result is concatenated with the corresponding encoder feature \((B,C,T,H,W)\) to form \((B,2C,T,H,W)\), which is passed through DoubleConv3D to yield \((B,C,T,H,W)\).

\subsection{MLP block}
The MLP block (Fig.~\ref{fig:block}c) embeds the diffusion timestep \(t\) into a higher-dimensional representation. It comprises a linear layer, SiLU activation, and a second linear layer, mapping the sinusoidal embedding to a 512-dimensional time feature.

\subsection{Overall computation detail in U-Net}
The full U-Net (Figure 2 in main paper) processes input \(x\) and condition \(c\) in parallel encoder-decoder streams. At each encoder level, the condition encoder and time MLP projections are added to the DoubleConv3D output before pooling. In the decoder, after upsampling and skip concatenation, the same condition and time features are added before the next DoubleConv3D. A final \(1\times1\times1\) convolution projects the last decoder output to the predicted velocity. U-Net structure is detailed in Algorithm~\ref{alg:unet_v_forward}.
\begin{algorithm}[htbp]
\caption{Forward pass of the U-Net \(v_{\theta}\)}
\label{alg:unet_v_forward}
\begin{algorithmic}[1]
  \Require 
    \(x_{t}\in\mathbb{R}^{B\times1\times T\times H\times W}\),\quad
    \(t\in\{1,\dots,T\}^{B}\),\quad
    \(c\in\mathbb{R}^{B\times C_{\mathrm{cond}}\times T\times H\times W}\)
  \Ensure 
    \(\hat{v}\in\mathbb{R}^{B\times1\times T\times H\times W}\)
  \State \Comment{1. Condition dropout}
  \If{training and \(p_{\mathrm{drop}}>0\)}
    \State Randomly set one channel of \(c\) to \(-1\) with probability \(p_{\mathrm{drop}}\)
  \EndIf
  \State \Comment{2. Time embedding}
  \State \(g \gets \mathrm{TimeMLP}(\mathrm{SinEmbed}(t))\)
  \State \Comment{3. Build condition features}
  \State \(z_{1}\gets \mathrm{CondInc}(c)\)
  \State \(z_{2}\gets \mathrm{CondD1}(z_{1})\)
  \State \(z_{3}\gets \mathrm{CondD2}(z_{2})\)
  \State \(z_{4}\gets \mathrm{CondD3}(z_{3})\)
  \State \Comment{4. Encoder path}
  \State \(x_{1}\gets \mathrm{Inc}(x_{t}) + \mathrm{Proj}(g;\texttt{inc}) + z_{1}\)
  \State \(x_{2}\gets \mathrm{E1}(x_{1}) + \mathrm{Proj}(g;\texttt{E1}) + z_{2}\)
  \State \(x_{3}\gets \mathrm{E2}(x_{2}) + \mathrm{Proj}(g;\texttt{E2}) + z_{3}\)
  \State \(x_{4}\gets \mathrm{E3}(x_{3}) + \mathrm{Proj}(g;\texttt{E3}) + z_{4}\)
  \State \Comment{5. Decoder path}
  \State \(u_{1}\gets \mathrm{D1}(x_{4},\,x_{3}) + \mathrm{Proj}(g;\texttt{D1}) + z_{3}\)
  \State \(u_{2}\gets \mathrm{D2}(u_{1},\,x_{2}) + \mathrm{Proj}(g;\texttt{D2}) + z_{2}\)
  \State \(u_{3}\gets \mathrm{D3}(u_{2},\,x_{1}) + \mathrm{Proj}(g;\texttt{D3}) + z_{1}\)
  \State \Comment{6. Predict velocity}
  \State \(\hat{v}\gets \mathrm{Head}(u_{3})\)
  \State \Return \(\hat{v}\)
\end{algorithmic}
\end{algorithm}

\begin{table}[tb]
  \centering
  \caption{Hyper-parameter settings for the 3D U-Net architecture.}
  \label{tab:unet_hparams}
  \renewcommand{\arraystretch}{1.1}
  \begin{tabularx}{\textwidth}{l l X}
    \toprule
    \textbf{Parameter}                & \textbf{Value}                & \textbf{Description}                               \\
    \midrule
    Input channels                    & 1                             & Number of channels in noisy input \(x\).                       \\
    Condition channels             & 10                            & Number of channels in condition \(c\).                       \\
    Base channel size                 & 64                            & Channels in the initial conv block.                \\
    Channel multipliers               & [1,\,2,\,4,\,8]               & Relative to base across encoder levels.            \\
    Temporal kernel size \(t_k\)      & 3                             & Kernel extent in time dimension.                   \\
    Spatial kernel size               & \(3 \times 3\)                & Kernel extent in height \& width.                  \\
    Pool / upsample stride            & (1,\,2,\,2)                   & Halves H,W per pool; time unchanged.               \\
    Activation function               & SiLU                          & Applied after each GroupNorm.                      \\
    Normalization                     & GroupNorm (4--32 groups)       & Groups = \(\max(4,\min(32,\lfloor C/4\rfloor))\). \\
    Dropout3d probability             & 0.0                           & Applied after each activation.                    \\
    SE reduction ratio \(r\)          & 16                            & Squeeze-and-Excitation bottleneck ratio.           \\
    Time embedding dim                & 128                           & Input dim for sinusoidal embedding MLP.            \\
    Time MLP hidden size              & 512                           & Two-layer MLP: 128→512→512.                        \\
    Time proj. dims (inc, d1, d2, d3) & (64,128,256,512)              & Linear proj's for encoder stages.                  \\
    Time proj. dims (u1, u2, u3)      & (256,128,64)                  & Linear proj's for decoder stages.                  \\
    Condition dropout \(p_{\rm drop}\)        & 0.2                           & Random channel drop in condition \(c\) during training.      \\
    Final output channels             & 1                             & Single-channel prediction output.                  \\
    \bottomrule
  \end{tabularx}
\end{table}

\section{Implementation Details of Diffusion Models}
\label{ap:diff}
In this section, we describe the implementation details of our diffusion models.

\subsection{Noise schedule}
The choice of noise schedule affects performance, as it governs how noise is injected throughout the diffusion process. In our study, we focus on two common schedules: the linear schedule and the cosine schedule.

\subsubsection{Linear noise schedule.} 
The linear noise schedule is the simplest schedule used in DDPMs \cite{ho2020denoising}. In this schedule, the variance \(\beta_t\) increases linearly by interpolating between \(\beta_{\min}\) and \(\beta_{\max}\), as shown in equation~\eqref{eq:linear_noise}. The quantities \(\alpha_t\) and \(\bar{\alpha}_t\) are then defined in equations.~\eqref{eq:linear_noise_alpha} and \eqref{eq:linear_noise_bar_alpha}, respectively.

\begin{align}
\label{eq:linear_noise}
\beta_t &= \beta_{\min}
  + \frac{t}{T-1}\,\bigl(\beta_{\max}-\beta_{\min}\bigr),
  && t = 0,1,\dots,T-1,\\
\label{eq:linear_noise_alpha}
\alpha_t &= 1 - \beta_t,\\
\label{eq:linear_noise_bar_alpha}
\bar\alpha_0 &= 1, 
  &\qquad&
\bar\alpha_t = \prod_{i=1}^t \alpha_i,\quad t=1,2,\dots,T.
\end{align}

\subsubsection{Cosine noise schedule.} 
The linear noise schedule pushes the data to near-pure noise in the later steps of the diffusion process and applies noise more steeply in the early steps. As an improved alternative, Nichol and Dhariwal \cite{nichol2021improved} proposed the cosine noise schedule, which increases noise more gradually than the linear schedule, particularly during the middle phase of diffusion. The cosine schedule defines the per-step variance \(\beta_j\) in terms of the cumulative noise level \(\bar\alpha_j\) as in equation~\eqref{eq:cosine_noise}.

\begin{align}
t_j &= \frac{j}{T}, 
  &\qquad& j = 0,1,\dots,T, \nonumber \\
\tilde\alpha(t_j)
  &= \cos^2\!\Bigl(\frac{t_j + s}{1 + s}\,\frac{\pi}{2}\Bigr), \nonumber \\
\bar\alpha_j
  &= \frac{\tilde\alpha(t_j)}{\tilde\alpha(0)},
  &\qquad& j = 0,1,\dots,T,\nonumber \\
  \label{eq:cosine_noise}
\beta_j
  &= 1 - \frac{\bar\alpha_j}{\bar\alpha_{j-1}},
  &\qquad& j = 1,2,\dots,T.
\end{align}

\section{DDPM algorithms in our study}
We define our masking, training and evaluation algorithms as Algorithm \ref{alg:masked_gen}, \ref{alg:training_v} and \ref{alg:sampling_v}, respectively. We used latitude weight $w(\phi)$:
\begin{equation}
  \label{eq:lat_weight}
     w(\phi_h) = \varepsilon + (1 - \varepsilon)\frac{\cos(\phi_h)}{\displaystyle \mathbb{E}_{\phi}\!\bigl[\cos(\phi)\bigr]},
    \quad
    \mathbb{E}_{\phi}[\cos(\phi)] = \frac{1}{H}\sum_{h=1}^{H}\cos(\phi_h),
\end{equation}

\begin{algorithm}[tb]
\caption{Masked Data Generation}
\label{alg:masked_gen}
\begin{algorithmic}[1]
    \For{\(i=1,\ldots,N\)}
        \State Extract the binary satellite observation flag map \(m_i\) from the GSMaP dataset.
        \State Extract the corresponding precipitation map \(x_{0}^i\) from the ERA5 dataset.
        \State Compute the masked precipitation map:
        \[
        \bar{x}_{0}^i = m_i \cdot x_{0}^i + (-1)\cdot\bigl(1-m_i\bigr),
        \]
        where all operations are performed element-wise.
    \EndFor
    \State \textbf{return}\ \(\{ \bar{x}_{0}^i \}_{i=1}^{N}\).
\end{algorithmic}
\end{algorithm}

\begin{algorithm}[tb]
  \caption{Training}
  \label{alg:training_v}
  \begin{algorithmic}[1]
    \For{each epoch}
      \For{$i=1,\ldots,N$}
        \State Sample a precipitation map $x_{0}^i$ from ERA5.
        \State Sample binary mask $m_i$ from GSMaP.
        \State Generate $\bar x_{0}^i$ via Alg.~\ref{alg:masked_gen}.
        \State Sample condition $c_i$, noise $\epsilon_i\!\sim\!\mathcal{N}(0,I)$, time $t$.
        \State Compute
          $x_{t}^i = \sqrt{\bar\alpha_t}\,x_{0}^i
                     + \sqrt{1-\bar\alpha_t}\,\epsilon_i$
          \\
        \State $v_{t}^i = \sqrt{\bar\alpha_t}\,\epsilon_i
                     - \sqrt{1-\bar\alpha_t}\,x_{0}^i$
        \State $\bar c_i \gets \{\bar x_{0}^i,\;c_i\}$
        \State $\displaystyle \theta \leftarrow \theta \;-\;\eta \,\nabla_{\theta}\;\Big\|\;w(\phi)\,\odot\bigl(v_{t}^i \;-\; v_{\theta}(x_{t}^i,\bar c_i,t)\bigr)\Big\|_{2}^{2}$ where $w(\phi)$ is from equation~\eqref{eq:lat_weight}

      \EndFor
    \EndFor
  \end{algorithmic}
\end{algorithm}

\begin{algorithm}[tb]
  \caption{Sampling}
  \label{alg:sampling_v}
  \begin{algorithmic}[1]
    \Require trained $v_\theta$, $\bar x_0$, mask $m\in\{0,1\}^{L\times H\times W}$, cond.~$c$, schedule $\{\alpha_t,\bar\alpha_t,\beta_t\}$
    \State $\bar c \gets \{\bar x_0,\,c\}$
    \State $x_T \gets \bar x_0$;\quad set $[x_T]_p\sim\mathcal N(0,1)$ where $m_p=0$
    \For{$t=T,\dots,1$}
      \State $z\sim\mathcal N(0,I)$ if $t>1$, else $z\gets 0$
      \State $\hat v_t\gets v_\theta(x_t,\bar c,t)$
      \State $\hat\epsilon_t \gets \sqrt{1-\bar\alpha_t}\,x_t + \sqrt{\bar\alpha_t}\,\hat v_t$
      \State $x_{t-1}\!\gets\!\frac{1}{\sqrt{\alpha_t}}\Bigl(x_t - \frac{\beta_t}{\sqrt{1-\bar\alpha_t}}\hat\epsilon_t\Bigr) + \sqrt{\beta_t}\,z$
      \State enforce observed: $x_{t-1}\gets m\odot x_0 + (1-m)\odot x_{t-1}$
    \EndFor
    \State \Return $x_0$
  \end{algorithmic}
\end{algorithm}

\section{Algorithm Implementation Details}

\subsection{Common Settings}

We implemented all algorithms using the PyTorch library \cite{paszke2019pytorch} and uniformly applied the RAdamScheduleFree optimizer \cite{defazio2024road,schedule_free} across all experiments. All models were trained for 100 epochs, where each epoch processes the entire training dataset. For data loading, we set the number of workers to $\max(1, \text{num\_CPU} / 2)$ and used a batch size of $B = 8$.

Throughout development, we adopted the 3D U-Net architecture depicted in Figure 2 in main paper and Figure \ref{fig:block} as: (i) the denoising network for our DDPM variants, (ii) the decoder in the conditional VAE (cVAE), and (iii) the generator in the conditional GAN (cGAN). For both cVAE and cGAN models, we omitted all components related to diffusion time embedding --- namely the sinusoidal embedding, the associated MLP block, and the linear projection layers.

We maintained an exponential moving average (EMA) of the model parameters with decay factor $\gamma = 0.999$. At each training step, the EMA weights $\theta_{\text{EMA}}$ are updated via:
\begin{equation}
\theta_{\text{EMA}} \leftarrow \gamma \theta_{\text{EMA}} + (1 - \gamma) \theta_{\text{current}},
\end{equation}
and these smoothed parameters are used for both evaluation and saving model checkpoints. Mixed-precision training (AMP) with gradient scaling is employed across all methods.

\subsubsection{Data Augmentation}
During training, we apply on-the-fly augmentations to both the condition and the target:
\begin{itemize}
  \item \textbf{Longitude flip:} with probability $0.5$, reflect the data along the east--west axis.
  \item \textbf{Latitude flip:} with probability $0.5$, reflect the data along the north--south axis.
  \item \textbf{Rotation:} with probability $0.5$, rotate by $180^{\circ}$ in the horizontal plane.
\end{itemize}
These operations are performed independently for each training sample.

\subsection{DDPM (Our Method)}

The implementation of our DDPM-based approach is detailed in Section 2 in main paper and Section \ref{ap:diff}. We used a learning rate of $1 \times 10^{-4}$ and weight decay of $1 \times 10^{-4}$ with the RAdamScheduleFree optimizer.

\subsubsection{DDIM Sampler}
For evaluation, we employed the DDIM sampler \cite{song2020denoising} with 50 steps using trained DDPM model weights (linear noise schedule, $T = 1000$). The DDIM sampler was applied only during evaluation sampling and did not require additional training.

\subsection{Conditional VAE (cVAE)}

\subsubsection{Model Architecture}
The cVAE model consists of the following components:

\paragraph{Encoder}
\begin{itemize}
  \item \textbf{Input:} Concatenate masked input $\bar{x}$ and condition $\bar{c}$ along the channel dimension.
  \item \textbf{Downsampling:} Four stages of downsampling, where each stage applies a DoubleConv3D block (with channels 64, 128, 256, 512) followed by max-pooling with kernel/stride $(1,2,2)$.
\end{itemize}

\paragraph{Latent Projection}
\begin{equation}
\mu, \log\sigma^2 = \text{Conv3d}_{1 \times 1 \times 1}(512 \rightarrow C_z) \quad \text{with} \quad C_z = 128, \; (L', H', W') = (L, H/8, W/8).
\end{equation}

\paragraph{Reparameterization}
\begin{equation}
z = \mu + \exp\left(\frac{1}{2}\log\sigma^2\right) \odot \epsilon, \quad \epsilon \sim \mathcal{N}(0, I).
\end{equation}

\paragraph{Decoder}
\begin{enumerate}
  \item Compute the channel-wise average of $z$, yielding shape $(B, 1, L', H', W')$, and upsample via trilinear interpolation to $(B, 1, L, H, W)$, which produces $z_{\text{recon}}$.
  \item Pass $(z_{\text{recon}}, \bar{c})$ as the input and condition to the 3D U-Net to produce the reconstruction $\hat{x} \in \mathbb{R}^{B \times 1 \times L \times H \times W}$.
\end{enumerate}

\subsubsection{Objective Function}
We optimize the evidence lower bound (ELBO) with a $\beta$-VAE \cite{higgins2017beta} formulation:
\begin{equation}
\mathcal{L}_{\text{cVAE}} = \underbrace{\mathbb{E}\left[\|w(\phi) \odot (\hat{x} - x)\|_2^2\right]}_{\text{latitude-weighted MSE}} + \beta \underbrace{\mathbb{E}\left[\text{KL}\left(\mathcal{N}(\mu, \sigma^2) \| \mathcal{N}(0, I)\right)\right]}_{\text{KL divergence}},
\end{equation}
where $w(\phi)$ is defined in equation~\eqref{eq:lat_weight} and $\beta = 0.001$.

\subsubsection{Training Setup}
Optimizer: RAdamScheduleFree with learning rate $1 \times 10^{-4}$ and weight decay $1 \times 10^{-4}$.

\subsubsection{Evaluation}
At test time, we use only the decoder: we sample latent codes $z \sim \mathcal{N}(0, I)$, supply $z$ and the condition $\bar{c}$ as inputs to the decoder, and obtain the inpainted output $\hat{x}$.

\subsection{Conditional GAN (cGAN)}

\subsubsection{Model Architecture}
We train a conditional generative adversarial network following the pix2pix style \cite{isola2017image}, using our 3D U-Net as generator $G$ and a 3D PatchGAN as discriminator $D$. At each training step, the generator $G$ receives the masked map $\bar{x}$ and the full condition $\bar{c}$ and produces $\hat{x} = G(\bar{x}, \bar{c})$. The discriminator $D$ processes the real pair $(x, \bar{c})$ as real and the fake pair $(\hat{x}, \bar{c})$ as fake, outputting a "realness" score map of the same spatial dimensions.

\paragraph{Discriminator Architecture}
The discriminator is a 3D PatchGAN that maps an input tensor of shape $(B, C_{\text{in}}, L, H, W)$ to a single-channel score volume:

\begin{enumerate}
  \item \textbf{Input:} Concatenation of $C_{\text{cond}}$ condition channels and 1 generated (or real) channel, so $C_{\text{in}} = C_{\text{cond}} + 1$.
  
  \item \textbf{Layer 1:}
    \begin{itemize}
      \item 3D convolution: $\text{in\_channels} = C_{\text{in}}$, $\text{out\_channels} = n_{\text{df}} = 64$
      \item Kernel size $(1, 4, 4)$, stride $(1, 2, 2)$, padding $(0, 1, 1)$
      \item Spectral normalization and LeakyReLU activation (negative slope 0.2)
    \end{itemize}
  
  \item \textbf{Layers 2 to $N$ (default $N = 4$):}
    \begin{itemize}
      \item 3D convolution, doubling channels each time (capped at 512): $n_f = \min(2 n_{f_{\text{prev}}}, 512)$
      \item Kernel $(1, 4, 4)$, stride $(1, 2, 2)$, padding $(0, 1, 1)$
      \item Spectral normalization, GroupNorm with $G = \max(4, \min(32, n_f/4))$ groups
      \item LeakyReLU(0.2) and Dropout3d with $p = 0.2$
    \end{itemize}
  
  \item \textbf{Final Layer:}
    \begin{itemize}
      \item 3D convolution to 1 channel, kernel $(1, 4, 4)$, stride $(1, 1, 1)$, padding $(0, 1, 1)$
      \item Spectral normalization, no activation (raw logits output)
    \end{itemize}
\end{enumerate}

\subsubsection{Objective Function}
We adopt hinge-style adversarial losses \cite{lim2017geometric}. The discriminator loss is:
\begin{equation}
\mathcal{L}_D = \frac{1}{2}\mathbb{E}[\max(0, 1 - D(x, \bar{c}))] + \frac{1}{2}\mathbb{E}[\max(0, 1 + D(\hat{x}, \bar{c}))],
\end{equation}
and the generator adversarial loss is:
\begin{equation}
\mathcal{L}_G^{\text{adv}} = -\mathbb{E}[D(\hat{x}, \bar{c})].
\end{equation}

To encourage accurate reconstruction, we add a latitude-weighted $L_1$ loss:
\begin{equation}
\mathcal{L}_G^{L_1} = \mathbb{E}[\|w(\phi) \odot (\hat{x} - x)\|_1],
\end{equation}
where $w(\phi)$ is from equation~\eqref{eq:lat_weight} with $\varepsilon = 0.01$. The total generator loss is:
\begin{equation}
\mathcal{L}_G = \mathcal{L}_G^{\text{adv}} + \lambda_1 \mathcal{L}_G^{L_1}, \quad \lambda_1 = 1.0.
\end{equation}

\subsubsection{Training Setup}
\begin{itemize}
  \item Two-time-scale update rule (TTUR) \cite{heusel2017gans} with RAdamScheduleFree optimizers.
  \item Generator: learning rate $1 \times 10^{-4}$, weight decay $1 \times 10^{-4}$.
  \item Discriminator: learning rate $3 \times 10^{-4}$, no weight decay.
\end{itemize}

\subsection{Supervised Learning with U-Net}

\subsubsection{Model Architecture}
We implement a baseline supervised learning approach using the same 3D U-Net architecture as described in the common settings. The network directly maps from the concatenated input $[\bar{x}, \bar{c}]$ to the target output $x$, where $\bar{x}$ is the masked input and $\bar{c}$ is the condition.

\subsubsection{Objective Function}
The model is trained using a simple latitude-weighted $L_1$ loss:
\begin{equation}
\mathcal{L}_{\text{supervised}} = \mathbb{E}[\|w(\phi) \odot (\hat{x} - x)\|_1],
\end{equation}
where $\hat{x} = f_\theta([\bar{x}, \bar{c}])$ is the network prediction, $x$ is the ground truth, and $w(\phi)$ is the latitude weighting function defined in equation~\eqref{eq:lat_weight}.

\subsubsection{Training Setup}
Optimizer: RAdamScheduleFree with learning rate $1 \times 10^{-4}$ and weight decay $1 \times 10^{-4}$.

\subsection{Rectified Flow}

\subsubsection{Model Architecture}
Rectified Flow \cite{liu2022flow} is implemented following the velocity prediction paradigm. Let $x_1$ be the ground-truth target volume, $x_0 \sim \mathcal{N}(0, I)$ a Gaussian noise sample of the same shape, and $\bar{c}$ the condition.

\subsubsection{Training Procedure}
For each minibatch, we sample $s \sim \text{Uniform}(0, 1)$ and form the linear interpolation:
\begin{equation}
x_s = (1 - s) x_0 + s x_1.
\end{equation}
The target velocity is given by:
\begin{equation}
\frac{\partial x_s}{\partial s} = x_1 - x_0 = v_{\text{target}}.
\end{equation}

\subsubsection{Objective Function}
The model $v_\theta(x_s, s, \bar{c})$ is trained to predict the velocity using a latitude-weighted mean squared error loss:
\begin{equation}
\mathcal{L}(\theta) = \mathbb{E}_{s, x_0}\left[\|w(\phi) \odot (v_\theta(x_s, s, \bar{c}) - (x_1 - x_0))\|_2^2\right],
\end{equation}
where $w(\phi)$ is the latitude weighting function defined in equation~\eqref{eq:lat_weight}.

\subsubsection{Training Setup}
The implementation follows the same common settings described above, using RAdamScheduleFree optimizer with learning rate $1 \times 10^{-4}$ and weight decay $1 \times 10^{-4}$.

\section{Evaluation Metrics}

Let \(X_{fij}\) and \(\hat X_{fij}\) denote the ground-truth and predicted precipitation at time frame \(f\in\{1,\dots,L\}\) and spatial location \((i,j)\).  We introduce the \emph{hole mask}
\[
M_{fij} \;=\;
\begin{cases}
1,&\text{if the model was required to inpaint at }(f,i,j),\\
0,&\text{otherwise},
\end{cases}
\]
and define the hole domain \(H=\{(f,i,j)\mid M_{fij}=1\}\) of size \(\lvert H\rvert\).  All metrics defined above are computed over the hole domain \(H\), with the exception of the Boundary Discontinuity Index (BDI), which is computed over the one-pixel boundary \(B\).

\subsection{Root Mean Squared Error (RMSE)}  
Root Mean Squared Error quantifies the average magnitude of the inpainting error:
\[
\mathrm{RMSE}_{H}
=\sqrt{\frac{1}{\lvert H\rvert}\sum_{(f,i,j)\in H}\bigl(\hat X_{fij}-X_{fij}\bigr)^{2}}.
\]

\subsection{Temporal-Gradient RMSE (TG-RMSE)}  
TG-RMSE evaluates the accuracy of predicted temporal changes inside the hole.  Define finite differences
\(\Delta_{t}X_{fij}=X_{fij}-X_{(f-1)\,i\,j}\) for \(f=2,\dots,L\), and let \(H'=\{(f,i,j)\in H\mid f\ge2\}\).  Then
\[
\mathrm{TG\text{-}RMSE}_{H}
=\sqrt{\frac{1}{\lvert H'\rvert}\sum_{(f,i,j)\in H'}\bigl[\Delta_{t}\hat X_{fij}-\Delta_{t}X_{fij}\bigr]^{2}}.
\]

\subsection{Pearson Correlation Coefficient}  
Pearson's \(\rho\) measures linear agreement between \(\hat X\) and \(X\) over the hole:
\[
\rho_{H}
=\frac{\displaystyle\sum_{(f,i,j)\in H}
(\hat X_{fij}-\overline{\hat X}_{H})(X_{fij}-\overline{X}_{H})}
{\sqrt{\displaystyle\sum_{(f,i,j)\in H}(\hat X_{fij}-\overline{\hat X}_{H})^{2}}\;
\sqrt{\displaystyle\sum_{(f,i,j)\in H}(X_{fij}-\overline{X}_{H})^{2}}},
\]
where \(\overline{\hat X}_{H}\) and \(\overline{X}_{H}\) are means over \(H\).

\subsection{Multi-scale Structural Similarity (MS-SSIM)}  
For each frame \(f\), we compute 2D SSIM on the hole pixels:
\[
\mathrm{SSIM}_{f}
=\frac{(2\,\mu_{\hat X^{(f)}}\,\mu_{X^{(f)}}+C_{1})\,(2\,\sigma_{\hat X^{(f)}X^{(f)}}+C_{2})}
{(\mu_{\hat X^{(f)}}^{2}+\mu_{X^{(f)}}^{2}+C_{1})\,(\sigma_{\hat X^{(f)}}^{2}+\sigma_{X^{(f)}}^{2}+C_{2})},
\]
where all statistics use only \(\{(i,j)\mid M_{fij}=1\}\).  We then down-sample \(S\) times and average:
\[
\mathrm{MS\text{-}SSIM}_{f}
=\frac{1}{S}\sum_{s=1}^{S}
\mathrm{SSIM}\bigl(\downarrow^{s-1}\!\hat X^{(f)},\,\downarrow^{s-1}\!X^{(f)}\bigr),
\]
and finally average across frames:
\[
\mathrm{MS\text{-}SSIM}_{3D}
=\frac{1}{L}\sum_{f=1}^{L}\mathrm{MS\text{-}SSIM}_{f}.
\]

MS-SSIM was introduced by \cite{wang2003multiscale}.

\subsection{Boundary Discontinuity Index (BDI)}  
BDI measures the mean absolute difference along the one-pixel boundary \(B\) of the hole in the first frame:
\[
\mathrm{BDI}
=\frac{1}{\lvert B\rvert}\sum_{(i,j)\in B}\bigl|\hat X_{1ij}-X_{1ij}\bigr|,
\]
where \(B\) is obtained by dilating the hole by one pixel and taking the intersection with its complement.

%
\bibliography{jgr.bib} 
%

\end{document}